%% file: main.tex
\documentclass[sigplan,twocolumn]{acmart}
\usepackage{algorithm}
\usepackage{algpseudocode}
\algtext*{EndWhile}
\algtext*{EndIf}
\algtext*{EndFor}
\algtext*{EndProcedure}
\usepackage{amsmath}
\usepackage{amssymb}
\usepackage{array}
\usepackage{booktabs}
\usepackage{color}
\usepackage{comment}
\usepackage{cuted}
\usepackage[inline]{enumitem}
\usepackage{epsfig}
\usepackage{etoolbox}
\usepackage{fancybox}
\usepackage{float}
\usepackage{graphicx}
\usepackage{hyperref}
\usepackage{booktabs}
\usepackage{listings}
\usepackage{lipsum}
\usepackage{multirow}
\usepackage{multicol}
\usepackage{makecell}
\usepackage{placeins}
\usepackage{pgffor}
\usepackage{rotating}
\usepackage{setspace}
\usepackage[hang, tight]{subfigure}
\usepackage{threeparttable}
\usepackage{url}
\usepackage{verbatim}
\usepackage{wrapfig}
\usepackage[dvipsnames]{xcolor}

\usepackage{iftex}
\ifPDFTeX
  \usepackage[UTF8,scheme=plain,fontset=none]{ctex}
\else
  \usepackage[UTF8,scheme=plain,fontset=fandol]{ctex}
\fi
\makeatletter
\patchcmd{\paragraph}{-.5\baselineskip \@plus -2\p@ \@minus -.2\p@}{\z@}%
  {\typeout{ACMPAR: paragraph gap patched}}{\typeout{ACMPAR: paragraph gap patch FAILED}}
\let\ACM@origparagraph\paragraph
\makeatother
\makeatletter
\def\codeid@loop#1_#2\codeid@end{\ifx\relax#2\relax\mbox{#1}\else\mbox{#1}\_\allowbreak\codeid@loop#2\codeid@end\fi}
\newcommand{\codeid}[1]{\texttt{\codeid@loop#1_\codeid@end}}
\makeatother

\newbool{inccomment}
\setbool{inccomment}{true}
\graphicspath{{_fig/}}

\algrenewcommand\algorithmicindent{1.0em}%

\newcommand{\XC}[1]{\ifbool{inccomment}{{\color{magenta}YC\@: #1}}{}}
\newcommand{\ML}[1]{\ifbool{inccomment}{{\color{blue}XC\@: #1}}{}}
\newcommand{\TD}[1]{\ifbool{inccomment}{{\color{orange}#1}}{}}
\newcommand{\FN}[1]{\ifbool{inccomment}{{\color{OliveGreen}#1}}{}}

\newcommand{\ct}{\ifbool{inccomment}{{\color{magenta}$[$C$]$}}}
\newcommand{\llnn}{\ifbool{inccomment}{{\color{magenta}\\=============================================\\}}}

\newcommand{\textplaceholder}[3][gray]{%
  {%
    \color{#1}%
    \def\word{lorem}% 默认单词
    \if\relax\detokenize{#3}\relax\else\def\word{#3}\fi
    \foreach \i in {1,...,#2}{\word\ }%
  }%
}

\newcommand{\ours}{\textit{dit.xpu }}

\title{\ours: Taming DiT Sparse Inference on Mobile SoCs}

\author{Jiayu Chen}
\authornote{Jiayu Chen and Jinye Hu contributed equally to this work.}
\email{jiayu.chen.25@stu.pku.edu.cn}
\affiliation{%
  \institution{Peking University}
  \department{School of Computer Science}
  \city{Beijing}
  \country{China}
}

\author{Jinye Hu}
\authornotemark[1]
\email{2400013532@stu.pku.edu.cn}
\affiliation{%
  \institution{Peking University}
  \department{School of Electronics Engineering and Computer Science}
  \city{Beijing}
  \country{China}
}

\author{Maoliang Li}
\email{maoliang.li@stu.pku.edu.cn}
\affiliation{%
  \institution{Peking University}
  \department{School of Computer Science}
  \city{Beijing}
  \country{China}
}

\author{Anqian Cen}
\email{1120250956@bit.edu.cn}
\affiliation{%
  \institution{Beijing Institute of Technology}
  \department{School of Computer Science \& Technology}
  \city{Beijing}
  \country{China}
}

\author{Hailong Zou}
\email{zouhailong26@stu.pku.edu.cn}
\affiliation{%
  \institution{Peking University}
  \department{School of Computer Science}
  \city{Beijing}
  \country{China}
}

\author{Zihao Zheng}
\email{zhengzihao@stu.pku.edu.cn}
\affiliation{%
  \institution{Peking University}
  \department{School of Computer Science}
  \city{Beijing}
  \country{China}
}

\author{Xinming Wei}
\email{weixinming@pku.edu.cn}
\affiliation{%
  \institution{Peking University}
  \department{School of Computer Science}
  \city{Beijing}
  \country{China}
}

\author{Guojie Luo}
\email{gluo@pku.edu.cn}
\affiliation{%
  \institution{Peking University}
  \city{Beijing}
  \country{China}
}

\author{Jie Zhang}
\email{jiez@pku.edu.cn}
\affiliation{%
  \institution{Peking University}
  \city{Beijing}
  \country{China}
}

\author{Xuanzhe Liu}
\email{liuxuanzhe@pku.edu.cn}
\affiliation{%
  \institution{Peking University}
  \city{Beijing}
  \country{China}
}

\author{Xiang Chen}
\authornote{Corresponding author.}
\email{xiang.chen@pku.edu.cn}
\affiliation{%
  \institution{Peking University}
  \city{Beijing}
  \country{China}
}

\begin{document}

\input{_txt/0_abstract}

\maketitle

\renewcommand{\shortauthors}{Chen et al.}

% \begingroup
% \renewcommand\thefootnote{}
% \footnotetext{$^{*}$ Both authors contributed equally to this research}
% \footnotetext{$^{\dagger}$ Corresponding author: Xiang Chen, xiang.chen@pku.edu.cn}
% \endgroup

%=======================================%
%               Sections                %
%=======================================%
\input{_txt/1_introduction}

\input{_txt/2_background}

\input{_txt/3_motivation}

\input{_txt/4_optimization}
\input{_txt/5_implementation}
\input{_txt/6_experiment}

\input{_txt/7_related_work}
\input{_txt/8_conclusion}

%=======================================%
%=======================================%
% \clearpage
\balance
\nocite{*}
\bibliographystyle{ACM-Reference-Format}
\bibliography{
    _ref/1_dit_accel,
    _ref/2_dataset,
    _ref/3_mobile
}
%=======================================%
%=======================================%
\end{document}

%% file: _txt/0_abstract.tex
\begin{abstract}
Diffusion Transformers (DiTs) enable on-device image generation and editing but incur high attention costs at high resolutions. Existing sparse attention reduces computation, yet dynamic sparsity introduces preprocessing and irregular data accesses that poorly fit mobile NPUs. We present DiT.XPU, a sparse DiT inference framework for heterogeneous mobile SoCs. DiT.XPU combines tile-aligned sparse attention with anchor-based error compensation by exploiting slowly varying dense--sparse output residuals across denoising steps. Heterogeneous XPU collaboration overlaps NPU dense attention with GPU sparse calibration, while mask-guided dataflow reduces data transfers and layout conversion by overlapping data movement with computation. Across five DiT models on two Snapdragon smartphones, DiT.XPU achieves up to $1.56\times$ end-to-end denoising speedup over the Hexagon-NPU backend and $11.5\times$ attention-kernel speedup over the fastest alternative, while preserving generation quality close to dense attention. Code is available at \url{https://github.com/IF-LAB-PKU/dit.xpu}.
\end{abstract}

%% file: _txt/1_introduction.tex
\section{Introduction}
\label{sec:introduction}

Diffusion Transformer (DiT) architectures~\cite{dit,zimage,sd3} have enabled remarkable image generation capabilities with broad applications in digital content creation, image editing~\cite{flux2}, and visual design.  
    Yet these capabilities are not widely accessible through on-device execution on mobile devices, a pervasive platform for multimodal interaction, because their limited compute resources struggle to support the substantial demands of DiT inference.

% Background: why DiT inference is computationally demanding.
To understand this barrier, consider how DiTs perform inference. 
    Unlike autoregressive LLMs~\cite{liu2024deepseek,bai2023qwen}, which process new tokens using a KV cache, DiTs recompute attention over the entire visual token grid at every denoising step. 
    As resolution increases, the visual token count grows, and global attention cost scales quadratically with it. 
    This cost recurs throughout denoising, making inference compute-intensive. 
As shown in Fig~\ref{fig:attention-share}, Stable Diffusion 3.5 Medium requires nearly half an hour of denoising to generate a single $2048 \times 2048$ image on the Snapdragon 8Gen4 NPU, with attention accounting for 86\% of the latency.

% Background: dynamic sparse attention and its existing execution setting.
Sparse token interactions in DiT attention offer an opportunity to reduce computation. 
    Sparse attention retains important interactions while omitting those with little effect on the output~\cite{sparsetransformer,longformer,bigbird,reformer,minference}. 
Because their importance varies with denoising state, these methods dynamically map selected token-block computations onto hardware~\cite{spargeattn,pisa}. 
    This mapping introduces preprocessing overhead and irregular memory accesses, so existing approaches use custom CUDA kernels to exploit GPU parallelism. 
For example, Sparge\-Attn~\cite{spargeattn} uses fused CUDA kernels to reduce the combined preprocessing and mapping overhead.

% Background: mobile hardware constraints.
However, mobile SoCs struggle to efficiently execute DiT inference, even with sparse attention. 
    Compared with NVIDIA GPUs, mobile SoCs offer limited compute resources fragmented across heterogeneous CPUs, GPUs, and NPUs. 
    Even the NPU, their most capable processor for neural workloads, faces restrictive tensor layout requirements and limited on-chip memory~\cite{hexagonnpu}. 
Supporting sparse attention under these constraints requires layout conversion and carefully scheduled memory transfers, adding preprocessing and mapping overhead. 
    As shown in Fig.\ref{fig:dynamic_sparse_latency}, SpargeAttn spends 47.4\% of its total attention latency mapping the sparsity pattern.

% Challenge: three progressively coupled difficulties.
Based on our analysis of DiT inference mechanisms and mobile SoC constraints, we identify three progressively more complex challenges.

\textbf{(1) Heterogeneous XPU.} Given the limited capacity of mobile PUs, collaboration among heterogeneous cores has become mainstream, commonly termed the XPU computing paradigm. 
    These XPUs differ in computational capability and available resources, so DiT tasks must be assigned based on their execution characteristics. 
This requires partitioning the inference graph into tasks suited to different XPUs and arranging their execution as a pipeline. 
    The resulting pipeline must preserve task dependencies while efficiently exchanging intermediate tensors between XPUs.

\textbf{(2) Fragmented Dataflow.} Although XPU collaboration alleviates workload capacity constraints, fragmented computing resources further complicate dataflow. A partition that balances computation may still create mismatches in memory capacity and data layout. To fit within an XPU's local memory, a task may need to be split into smaller blocks, leading to repeated transfers of intermediate tensors. When tensors move between XPUs, they may need to be rearranged to match the receiving unit's data layout. Such overhead can erode the gains from XPU collaboration.

\textbf{(3) Approximation Errors.} Even after the workload is deployed, adapting the algorithm to hardware constraints can introduce approximation errors that threaten generation quality. For sparse attention, mapping an adaptive interaction pattern onto fixed-granularity computation limits its ability to track changes in interaction importance across denoising steps. Interactions omitted by the pattern may become important later, introducing approximation errors that can degrade quality. Compensating for these errors is necessary to retain quality under efficient XPU execution.

% Observation: the basis for reconciling regular execution and error control.
\textbf{To address these challenges, we present \ours, a training-free framework that leverages mobile XPU resources for efficient sparse DiT inference.} \ours introduces an approximate tile-aligned attention scheme that provides flexibility for hardware mapping through predictable computation and data access. The framework incorporates the following key designs:

\noindent\textbf{Hardware-Aligned Sparse Attention Algorithm.}
To characterize the approximation errors caused by mapping an adaptive interaction pattern onto fixed-granularity computation, we compare sparse and dense attention outputs across denoising steps. Although omitted interactions may become important later, our analysis shows that the output residual changes relatively slowly between adjacent steps. Accordingly, \ours computes both attention outputs on the same inputs at anchor steps and caches their difference as a calibration residual. The dense output serves the current step, while subsequent sparse steps reuse the residual to compensate for approximation errors, helping retain quality.

\noindent\textbf{Heterogeneous XPU Collaboration.}
To mitigate the latency overhead of anchor-based compensation, \ours orchestrates heterogeneous XPU collaboration by partitioning the anchor-step inference graph according to task characteristics and available resources: the NPU computes dense attention for the current step, while the GPU concurrently computes sparse attention for calibration. Asynchronous submission overlaps the two branches within a cross-XPU pipeline, while shared memory mappings give both processors direct access to intermediate tensors without explicit cross-processor copies. The runtime synchronizes the branches when updating the calibration residual or reusing buffers, preserving task dependencies while limiting coordination overhead.

\noindent\textbf{Mask-Guided Dataflow Optimization.}
To mitigate the transfer and rearrangement overhead caused by fragmented XPU resources, \ours derives a dataflow schedule from the static attention mask. By grouping contiguous accesses and reusing loaded blocks across neighboring computations, the schedule reduces repeated transfers when attention is partitioned to fit local memory. \ours further overlaps data movement with computation and produces intermediate tensors directly in the layout required by attention execution, avoiding separate rearrangement stages. Together, these mechanisms reduce the dataflow cost of XPU collaboration.

\textbf{Implementation and evaluation.}
We implement \ours in over 5,000 lines of C++ and Python and evaluate five DiT models on the OnePlus Ace 5 and OnePlus 13 at resolutions from $256 \times 256$ to $1024 \times 1024$. \ours achieves up to an $11.0\times$ attention-kernel speedup over the fastest alternative and a $1.56\times$ end-to-end denoising speedup over Stable-Diffusion.cpp's~\cite{sdcpp} Hexagon-NPU backend. Across the five models, GenEval improves by 0.54 percentage points on average, while HPSv2.1 and ImageReward decrease by only 0.06 and 0.01, respectively, relative to dense attention.

Our contributions are summarized as follows:
\begin{itemize}
    \item We analyze sparse DiT inference on mobile SoCs and identify three challenges: heterogeneous XPU execution, fragmented dataflow, and approximation errors.
    \item We present three techniques: anchor-based error compensation, heterogeneous XPU collaboration, and mask-guided dataflow optimization for tile-aligned sparse attention.
    \item We design and implement \ours, the first DiT sparse inference framework that coordinates mobile XPUs through asynchronous execution and shared memory.
    \item We demonstrate that, across five DiT models on two mobile devices, \ours achieves up to an $11.0\times$ attention-kernel speedup over the fastest alternative and a $1.56\times$ end-to-end denoising speedup over Stable-Diffusion.cpp's Hexagon NPU backend, while maintaining generation quality close to dense attention.
\end{itemize}

%% file: _txt/2_background.tex
\begin{figure}[t]
  \centering
  \includegraphics[width=\columnwidth]{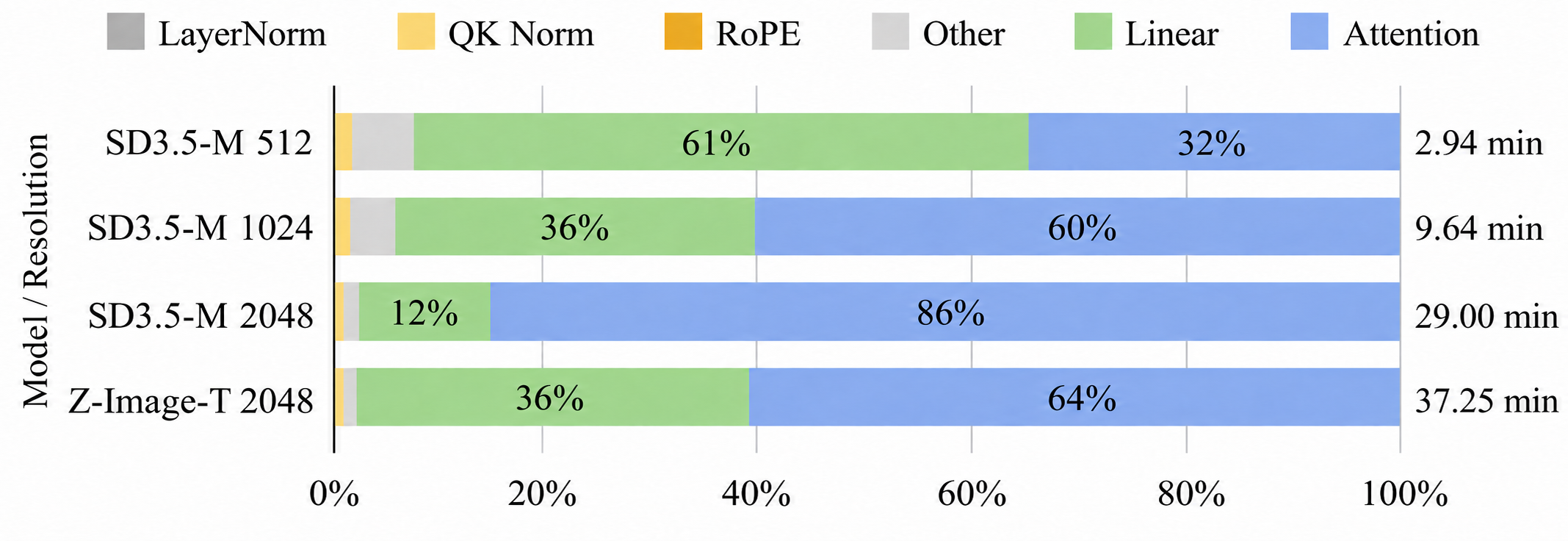}
  \caption{Attention dominates high-resolution DiT latency. It accounts for 32\%, 60\%, and 86\% of the SD3.5 Medium DiT runtime at 512, 1024, and 2048 resolution, respectively, and 64\% for Z-Image Turbo at 2048. }
  \label{fig:attention-share}
\end{figure}

\section{\textbf{Background}}
\label{sec:background}

\subsection{Inference Mechanism of DiTs}
\label{sec:bg-sparse-attention}

% the figure and to the attention list below when it is.
A DiT~\cite{dit} repeatedly applies its denoising network to the full visual token grid. For $N$ visual tokens, feed-forward layers and projections scale linearly with $N$, whereas self-attention scales quadratically. With $8\times$ latent downsampling and $2\times2$ patches, a $1024\times1024$ image contains 4096 visual tokens. As Figure~\ref{fig:attention-share} shows, attention's share of denoising latency in SD3.5 Medium~\cite{sd3} rises from $32\%$ at $512\times512$ to $86\%$ at $2048\times2048$. Reducing attention computation is therefore key to accelerating high-resolution denoising.

\begin{figure}[t]
  \centering
  \includegraphics[width=\columnwidth]{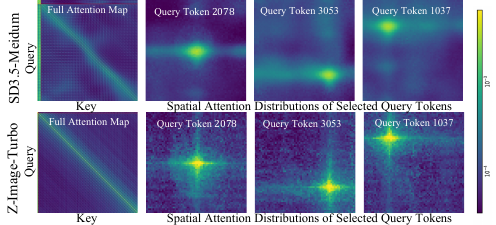}
  \caption{Attention patterns across DiT models. The first column shows the full Q–K attention map of a representative layer, while the remaining columns visualize the spatial attention distributions of selected query tokens.}
  \label{fig:attention-map}
\end{figure}

\noindent\textbf{DiT attention exhibits sparsity.}
Dense attention computes every query--key interaction, although their weights are highly uneven. Figure~\ref{fig:attention-map} visualizes this imbalance: its first column shows the full $Q$--$K$ attention map of a single layer, dominated by a narrow diagonal band, with only a few off-diagonal streaks reaching comparable strength; the last three columns show the spatial attention distributions of three selected query tokens over the spatial key grid. In both models the selected queries concentrate their weight on a compact, spatially local region and leave the rest of the grid nearly dark. This concentration motivates retaining influential interactions and skipping weaker ones~\cite{sparsevideogen,radialattn}.

\begin{figure}[t]
  \centering
  \includegraphics[width=\columnwidth]{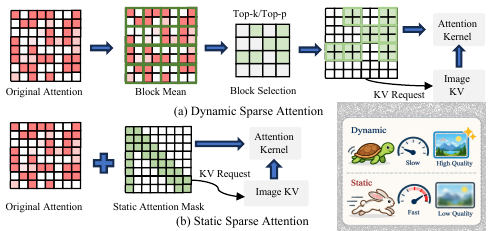}
  \caption{(a) Dynamic sparse attention workflow. (b) Static sparse attention
  workflow. The image K/V blocks are requested from the dynamically selected
  blocks in (a), and from the fixed mask in (b).}
  \label{fig:attention}
\end{figure}

\noindent\textbf{Sparse attention workflow.}
Sparse attention retains selected key blocks for each query block~\cite{spargeattn,xattention}. In \emph{dynamic sparse attention} (Figure~\ref{fig:attention}(a)), the query $Q$ and key $K$ are partitioned into blocks $Q_i$ and $K_j$. Averaging the tokens in each block yields representatives $\bar q_i$ and $\bar k_j$, whose similarity provides a coarse block score $s_{ij}$. High-scoring block pairs form a binary mask $M$, where $M_{ij}=1$ denotes a retained pair. Token-level attention is then computed only over selected blocks as $O_i=\operatorname{softmax}\!\left(\frac{Q_iK[M_i]^\top}{\sqrt d}\right)V[M_i]$, where $K[M_i]$ and $V[M_i]$ contain the key and value blocks selected for $Q_i$. Recomputing the block means, scores, and mask at each step allows the selected interactions to change with the attention pattern.

In \emph{static sparse attention} (Figure~\ref{fig:attention}(b)), $M$ is defined before sampling and reused across steps~\cite{diffastattn}. The same attention computation applies, but block averaging, scoring, and selection are skipped during denoising. This removes repeated selection while leaving interactions outside $M$ unavailable.

\subsection{Hardware Constraints of Mobile SoCs}
\label{sec:bg-mobile-soc}

\begin{figure}[t]
  \centering
  \includegraphics[width=\columnwidth]{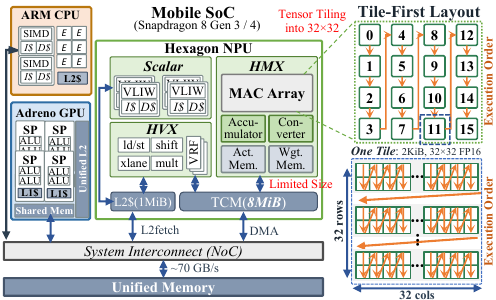}
  \caption{Architecture of a mobile SoC. Left: the CPU, GPU, and Hexagon NPU
  with their compute and memory resources. Right: the tensor stored in
  tile-first order, and the storage order within a $32\times32$ tile.}
  \label{fig:soc-architecture}
\end{figure}

\textbf{Fragmented XPU Resources in a Shared-Memory SoC.} Compute capacity in a mobile SoC is distributed across complementary processing units: an ARM CPU cluster, an Adreno GPU, and a Hexagon NPU~\cite{hexagonnpu}, as shown in Figure~\ref{fig:soc-architecture}. The CPU handles control flow, graph scheduling, and operators without dedicated kernels; the GPU handles workloads with irregular access patterns; and the NPU provides high throughput for dense matrix multiplication. All three share physical DRAM and its bandwidth through the system interconnect. Because DiT inference combines dense computation, irregular operations, and control tasks, efficient execution requires coordination across these processing units.

\textbf{Limited On-Chip Storage and Operand Staging.} HMX operands must be staged in limited on-chip memory~\cite{hexagonnpu}. The left side of Figure~\ref{fig:soc-architecture} shows the NPU's 1~MiB L2 cache and 8~MiB TCM, together with the activation and weight memories inside HMX. Part of the TCM can be configured as VTCM to supply data to HMX. The NPU moves data from DRAM through the L2fetch path and an asynchronous User\-DMA engine. When operands cannot fit in VTCM at once, they must be loaded in batches; insufficient reuse between batches causes repeated transfers.

\textbf{Tile-First Operand Layout and Conversion.} Within the Hexagon NPU, scalar cores execute control code, HVX vector cores perform SIMD operations, and the HMX matrix core performs tensor computation~\cite{hexagonnpu}. HMX accesses operands in tile-first order: in the upper-right illustration of Figure~\ref{fig:soc-architecture}, it visits tiles 0--3 before tiles 4--7, rather than traversing the entire matrix row by row. The tiles supplied to HMX must therefore be stored contiguously. CPUs and GPUs typically use row-major tensors, so passing their results to HMX requires layout conversion.

\textbf{Fixed Tile Granularity and Padding Overhead.} HMX performs matrix operations using fixed-size tiles~\cite{hexagonnpu}. For FP16 operands, each tile contains $32\times32$ elements and occupies 2~KiB, as illustrated on the right side of Figure~\ref{fig:soc-architecture}. When an operand dimension is not a multiple of 32, its boundary regions must be padded to full tiles.

%% file: _txt/3_motivation.tex
% Section fragment for a two-column EuroSys manuscript.
% Preamble requirement: \usepackage{booktabs}
% Replace all XX entries, device/model/resolution names, and citation keys
% with measured values and bibliography entries before submission.

% The outer border of each placeholder is exactly 3.3 in x 2 in.

\section{Motivation and Challenges}
\label{sec:motivation}

We first examine why dynamic sparse attention maps poorly to a mobile NPU.
We then identify an opportunity in the residual between static sparse and
dense attention, and characterize the hardware costs of exploiting it.

\subsection{Empirical Motivation}
\label{sec:dynamic_misalignment}

\begin{figure}[t]
  \centering
  \includegraphics[width=\columnwidth]{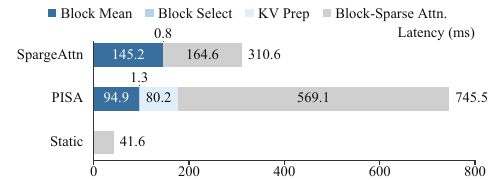}
  \caption{Dynamic and static sparse attention kernel breakdown for SD3.5 Medium at 1024*1024 on 8Gen4 NPU.}
  \label{fig:dynamic_sparse_latency}
\end{figure}

As discussed in Section~\ref{sec:bg-sparse-attention}, dynamic sparse attention must select and map token-block interactions before the attention kernel can run. This preparation lies on the critical path. As shown in Figure~\ref{fig:dynamic_sparse_latency}, at 40\% sparsity, SpargeAttn~\cite{spargeattn} spends 146.0\,ms on block-mean computation and block selection before its 164.6\,ms attention kernel. At 80\% sparsity, PISA~\cite{pisa} spends 176.4\,ms on block-mean computation, block selection, and K/V preparation before its 569.1\,ms kernel. Preparation thus accounts for 47.0\% and 23.7\% of their respective attention latencies, consuming a substantial part of the benefit from skipping token interactions.

The cost extends into kernel execution. Selected K/V blocks can differ across query blocks and denoising steps, so their transfers and VTCM staging cannot follow a schedule fixed in advance. This is consequential for Hexagon HMX, which requires operands in a compatible tile layout. At the same 80\% sparsity, PISA's block-sparse attention kernel takes 569.1\,ms, compared with 41.6\,ms for static block-sparse attention, even after preparation time is excluded. This kernel-level gap is consistent with the difficulty of executing dynamic block access efficiently on the mobile NPU. 

By contrast, static block-sparse attention fixes its interaction mask before inference. Guided by the attention pattern in Fig.~\ref{fig:attention-map} and prior local-window designs~\cite{longformer}, it retains a predefined diagonal band of token-block interactions. This fixed pattern removes per-step block selection and makes K/V accesses predictable.  This predictable execution is reflected in the 41.6\,ms attention-kernel latency shown in Fig.~\ref{fig:dynamic_sparse_latency}. Its fixed mask, however, cannot track changes in interaction importance across inputs and denoising steps, and may omit interactions needed for generation. Relative to dense attention, static sparsity reduces GenEval and HPSv2.1 by 41.54\% and 8.72\%, respectively, on Stable Diffusion 3.5 Medium. Across five DiT models, the mean relative reductions are 34.22\% and 16.61\%, respectively, as shown in Table~\ref{tab:static_quality}.

\subsection{Algorithmic Opportunity}
\label{sec:residual_opportunity}

\begin{figure}[t]
  \centering
  \includegraphics[width=\columnwidth]{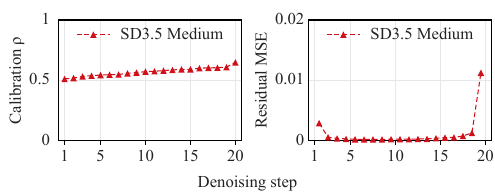}
  \caption{Calibration coefficient $\rho$ and residual MSE across denoising steps for SD3.5 Medium at $1024 \times 1024$ resolution.}
  \label{fig:residual_stability}
\end{figure}

Inspired by Re-ttention~\cite{reattention} and DiTFastAttn~\cite{diffastattn}, we examine dense and static sparse attention computed from the same $Q_t$, $K_t$, and $V_t$ at each denoising step $t$. Sparse attention omits some keys and normalizes the retained scores over a smaller set. It therefore removes the omitted keys' contributions while changing the weights assigned to retained keys. For query token $i$, we quantify this normalization shift with
\begin{equation}
\rho_{t,i}
=\exp\!\left(
\operatorname{LSE}_{t,i}^{s}
-\operatorname{LSE}_{t,i}^{d}
\right),
\end{equation}
where the log-sum-exp terms cover the retained and all keys, respectively. The coefficient $\rho_{t,i}$ is the fraction of dense attention probability retained by the static mask; multiplying the sparse output $O_{t,i}^{s}$ by $\rho_{t,i}$ restores the dense normalization of those retained interactions. Figure~\ref{fig:residual_stability}(a) shows that $\rho$ remains below one but changes gradually across denoising steps. The normalization shift is therefore substantial, while its coefficient varies smoothly.

We define the remaining residual as
\begin{equation}
R_{t,i}=O_{t,i}^{d}-\rho_{t,i}O_{t,i}^{s},
\end{equation}
which captures the omitted keys' contribution under dense attention. Figure~\ref{fig:residual_stability}(b) reports the MSE of this residual. It stays low for most denoising steps, although it rises near the beginning and end. Thus, after correcting the normalization of retained interactions, the measured output discrepancy is small through most of the denoising trajectory. Together, the gradual variation in $\rho$ and the low residual MSE motivate periodically computing these terms at anchor steps to compensate subsequent static sparse attention outputs.

\input{_tab/tab_static_loss}

\input{_tab/tab_dataflow_overhead}

\subsection{Deployment Challenges}
\label{sec:residual_challenges}

The temporal stability of the dense--sparse residual offers a way to
compensate for the approximation errors of static sparse attention.
Realizing this opportunity on a mobile SoC, however, requires addressing
two deployment challenges: coordinating calibration across XPUs and
feeding sparse computation from limited on-chip memory.

\paragraph{Heterogeneous XPU execution.}
As Fig.~\ref{fig:soc-architecture} shows, the CPU, GPU, and NPU
share unified memory through the system interconnect but differ in
execution engines and local storage. Compensating for static sparse
attention requires dense and sparse outputs computed from the same
$Q$, $K$, and $V$, since the sparse output alone cannot reveal the
approximation error. Computing both serially adds dense attention
work and may diminish the benefit of sparsity. Distributing them
across XPUs can overlap computation, but placement must account for
device capabilities and shared input access. The correction can be
formed only after both outputs are ready, so task submission, tensor
movement, and synchronization may offset that overlap; unified
physical memory alone does not guarantee compatible device mappings,
layouts, or buffer visibility.

\paragraph{Fragmented dataflow.}
The NPU in Fig.~\ref{fig:soc-architecture} has only 8\,MiB of
on-chip TCM, whereas one layer's image K/V in SD3.5 Medium occupies
24.0\,MiB. Q tiles, text K/V, intermediate results, and staging
buffers further reduce the space available for image K/V. The
simplest policy loads the blocks required by one query block,
computes its output, and releases that storage before moving to the
next query block. This limits the live working set, but neighboring
queries need overlapping image K/V blocks and repeatedly access text
K/V, causing the same data to be fetched again. As
Table~\ref{tab:kv_movement} shows, this policy reads 456.3\,MB from
DRAM. Cross-block reuse retains shared K/V blocks while neighboring
queries are processed and reduces reads to 216.4\,MB. However,
retained blocks compete with Q, text K/V, and staging buffers for
TCM space, limiting how far reuse and prefetching can extend. The
remaining DMA requests average only 5.32\,KB, and fetched blocks
still require conversion to HMX's tile-first layout. Consequently,
reuse alone leaves 2.11\,ms of data wait and 3.40\,ms of repacking.
The challenge is to coordinate block residency, query order, data
movement, and layout under the same limited TCM budget.

%% file: _tab/tab_static_loss.tex
\begin{table}[t]
  \caption{Static block-sparse attention substantially degrades generation
    quality across five DiT models.}
  \label{tab:static_quality}
  \centering
  \small
  \setlength{\tabcolsep}{3pt}
  \begin{tabular}{lrrrrrr}
    \toprule
    & \multicolumn{2}{c}{GenEval (\%) $\uparrow$} &
      \multicolumn{2}{c}{HPSv2.1 $\uparrow$} &
      \multicolumn{2}{c}{ImageReward $\uparrow$} \\
    \cmidrule(lr){2-3}\cmidrule(lr){4-5}\cmidrule(l){6-7}
    Model & Dense & Static & Dense & Static & Dense & Static \\
    \midrule
    SD3.5 Medium   & 68.37 & 26.83 & 29.18 & 20.46 & 0.8774 & $-0.7486$ \\
    SD3.5 Large    & 71.31 & 23.66 & 29.91 & 19.35 & 0.8781 & $-1.3234$ \\
    Z-Image        & 75.35 & 65.88 & 27.74 & 25.50 & 0.6936 & 0.2371 \\
    Z-Image-Turbo  & 75.49 & 63.14 & 29.83 & 28.59 & 0.6960 & 0.4741 \\
    FLUX.2 [klein] & 82.66 & 70.60 & 30.09 & 28.39 & 0.9108 & 0.5522 \\
    \bottomrule
  \end{tabular}
\end{table}

%% file: _tab/tab_dataflow_overhead.tex
\begin{table}[t]
  \caption{K/V data movement and kernel latency for SD3.5 Medium
  static sparse attention on the 8Gen4 NPU.}
  \label{tab:kv_movement}
  \centering
  \footnotesize
  \setlength{\tabcolsep}{1pt}
  \begin{tabular}{lccc}
    \toprule
    Measurement &
    \shortstack{Per-query loading} &
    \shortstack{Cross-query reuse} &
    \ours \\
    \midrule
    DRAM read volume (MB)      & 456.3 & 216.4 & 51.36 \\
    Mean DMA request size (KB) & 15.33 & 5.32  & 7.42  \\
    Data-wait time (ms)        & 4.52  & 2.11  & 0.02  \\
    K/V repacking time (ms)    & 5.45  & 3.40  & 0     \\
    Kernel latency (ms)        & 24.96 & 21.35 & 13.30 \\
    \bottomrule
  \end{tabular}
\end{table}

%% file: _txt/4_optimization.tex
\begin{figure}[t]
  \centering
  \includegraphics[width=\columnwidth]{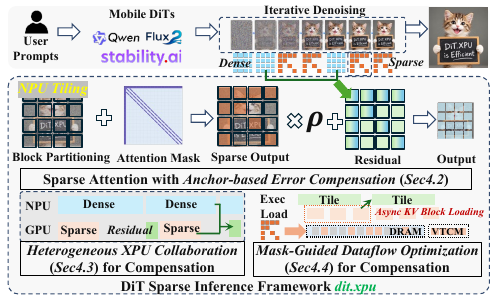}
  \caption{Overview of \ours.}
  \label{fig:overview}
\end{figure}

\section{System Design}
\label{sec:design}

\subsection{Design Overview}
\label{sec:design-overview}

As shown in Figure~\ref{fig:overview}, we present \ours, a training-free framework for sparse DiT inference on heterogeneous mobile SoCs. To translate reduced attention computation into end-to-end denoising speedup, its design addresses three challenges: approximation errors, heterogeneous XPU execution, and fragmented dataflow.

\paragraph{Hardware-Aligned Sparse Attention Algorithm.}
\ours uses a static, tile-aligned block-sparse attention mask so that retained interactions execute as regular matrix tiles on the NPU. The fixed mask avoids per-step pattern selection and makes computation and data access predictable. Because omitted interactions may become important during denoising, \ours periodically computes dense and sparse attention from the same Q/K/V at anchor steps. It caches their output difference as a calibration residual: the dense output serves the current step, while subsequent sparse steps reuse the residual to compensate for approximation errors.

\paragraph{Heterogeneous XPU Collaboration.}
At each anchor step, the NPU computes dense attention for the generation path while the GPU computes sparse attention for calibration. Asynchronous submission overlaps the two branches in a cross-XPU pipeline. Shared memory mappings let both processors access the same intermediate tensors without cross-processor copies. The runtime synchronizes the branches before updating the calibration residual, preserving task dependencies while limiting anchor-step overhead.

\paragraph{Mask-Guided Dataflow Optimization.}
The static mask specifies which K/V blocks each query block accesses, allowing \ours to plan transfers before execution. When the K/V working set exceeds VTCM capacity, the schedule groups contiguous transfers and reuses loaded blocks across neighboring query blocks. Double buffering overlaps loading the next blocks with NPU computation on the current blocks. Intermediate tensors are produced directly in the layout required by attention execution, avoiding separate rearrangement stages and reducing dataflow stalls.

\subsection{Hardware-Aligned Sparse Attention Algorithm}
\label{sec:4.2}

As image resolution increases, DiTs recompute global attention over a growing visual token grid at every denoising step, with cost quadratic in token count. Sparse attention reduces this cost, but mapping it to mobile NPUs presents two issues:
\begin{itemize}
    \item Fine-grained sparsity does not match fixed NPU matrix tiles, while dynamic block selection incurs per-step preprocessing and mapping overhead.
    \item A static mask makes computation and data access predictable, but omitted interactions introduce approximation errors through a normalization shift and missing key contributions.
\end{itemize}
We therefore combine a tile-aligned static mask with periodic anchor calibration. The mask executes retained interactions as regular matrix blocks; anchor steps compute the normalization coefficient and residual used to calibrate subsequent sparse outputs.

\paragraph{Tile-Aligned Static Halo Mask.}
Dynamic block-sparse attention first estimates the importance of query and K/V blocks, then selects K/V blocks for each query block using Top-$k$ or Top-$p$. Designed largely for NVIDIA GPUs, these methods use programmable CUDA kernels for block reduction and selection, then access selected K/V blocks through indices generated at runtime.

As discussed in Section~\ref{sec:motivation}, the Hexagon NPU's HMX executes fixed-layout matrix tiles in VTCM. Computing Q/K block means and performing Top-$k$/Top-$p$ selection require reductions, comparisons, and index generation, which rely mainly on the lower-throughput HVX. Even if HMX computes inter-block scores, block means must first be prepared and results exchanged between compute units. This preprocessing and mapping must finish before attention and recur at every layer and denoising step.

We avoid this repeated work by fixing, before inference, the image K/V blocks accessed by each image query block. We group 32 spatially adjacent image tokens into a block, so every retained query--K/V block pair forms a $32\times32$ attention matrix tile aligned with HMX's FP16 granularity. Because these block pairs are predetermined, the system can plan K/V transfer order and VTCM placement in advance, then follow that plan throughout denoising.

Specifically, image blocks form a two-dimensional grid. For image-to-image attention, a query block at $(y,x)$ retains K/V blocks satisfying $|y-y'|\leq1$ and $|x-x'|\leq1$, clipped at grid boundaries. Each query block therefore computes at most $3\times3$ image block pairs. Neighboring query blocks share much of their K/V working set, allowing the system to merge transfers, reuse loaded blocks in VTCM, and skip matrix tiles outside the halo.

\paragraph{Anchor-Based Error Compensation.}
A static mask can omit interactions that become important later in denoising, introducing approximation errors. It removes the omitted keys' contributions and changes the Softmax normalization of retained keys. As shown in \S~\ref{sec:residual_opportunity}, $\rho_{t,i}$ changes gradually across steps, while the reported MSE of $R_{t,i}$ stays low through most of the denoising trajectory. These observations motivate periodic anchor steps.

We designate an anchor step $a$ every $p$ denoising steps and compute dense and static sparse attention from the same $Q_a$, $K_a$, and $V_a$. The dense output $O_{a,i}^{d}$ serves the current step. For each query token $i$, we obtain $\rho_{a,i}$ by exponentiating the difference between $\operatorname{LSE}_{a,i}^{s}$ and $\operatorname{LSE}_{a,i}^{d}$, then compute the omitted-key residual $R_{a,i}$ as defined above. A subsequent sparse step $t$ reuses the terms from the most recent anchor:
\begin{equation}
    \widetilde{O}_{t,i}
    = \rho_{a,i}O_{t,i}^{s} + R_{a,i}.
\end{equation}
This approximately compensates for the normalization shift and omitted-key contributions without recomputing dense attention at every step.

\begin{figure}[t]
  \centering
  \includegraphics[width=\columnwidth]{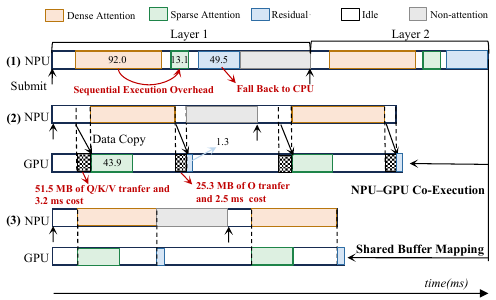}
  \caption{An illustration of the heterogeneous XPU collaboration.(1) sequential execution with CPU fallback, (2) NPU--GPU co-execution with explicit data copies, and (3) NPU--GPU co-execution through shared memory mappings.}
  \label{fig:pipeline}
\end{figure}

\subsection{Heterogeneous XPU Collaboration}
\label{sec:4.3}

At an anchor step $a$, the NPU dense branch and GPU sparse branch read the same $(Q_a,K_a,V_a)$ within each attention layer. The GPU uses the same static mask $M$ as subsequent NPU sparse steps. Both branches produce attention outputs and row-wise log-sum-exp values; once both results are ready, the runtime updates $(\rho_{a,i},R_{a,i})$ for each query $i$. The dense output $O_{a,i}^{d}$ serves the current step. A later sparse step $t$ computes $O_{t,i}^{s}$ on the NPU and applies the latest calibration terms to obtain $\widetilde{O}_{t,i}=\rho_{a,i}O_{t,i}^{s}+R_{a,i}$. Serializing the two anchor branches would place the auxiliary sparse computation on the generation path. \ours instead overlaps their execution and shares their intermediate tensors.

\paragraph{NPU--GPU Co-Execution.}
Once $(Q_a,K_a,V_a)$ is ready, the dense and sparse branches have no dependency on each other. The runtime submits dense attention to the NPU for the current step and static sparse attention to the GPU for calibration. After producing $O_{a,i}^{d}$, the NPU proceeds with the main path while the GPU completes its branch. The runtime waits for both outputs before publishing the new calibration terms; a subsequent sparse step reads only a fully published correction. It also keeps the input buffers available until the GPU has finished reading them. Figure~\ref{fig:pipeline}(1) and (2) illustrate how this schedule moves the auxiliary sparse branch out of serial NPU execution and overlaps it with the main path.

\paragraph{Shared Buffer Mapping.}
Co-execution would require copies at each anchor step if the NPU and GPU used separate tensor storage. Figure~\ref{fig:pipeline}(2) shows a 51.5~MB Q/K/V transfer costing 3.2~ms and a 25.3~MB output transfer costing 2.5~ms. To remove these transfers, \ours allocates Q/K/V, branch outputs, and calibration terms in shared physical memory, then maps the buffers into the CPU, NPU, and GPU address spaces. The GPU reads the NPU-produced Q/K/V from the shared buffers, and the calibration update reads both branch outputs without a staging copy. Device events ensure that readers see completed writes and that buffers are not reused while a branch still needs them. The runtime retains mappings across blocks to avoid repeated mapping setup. Figure~\ref{fig:pipeline}(3) illustrates the execution without copy stages.

\begin{figure}[t]
    \centering
    \includegraphics[width=\columnwidth]{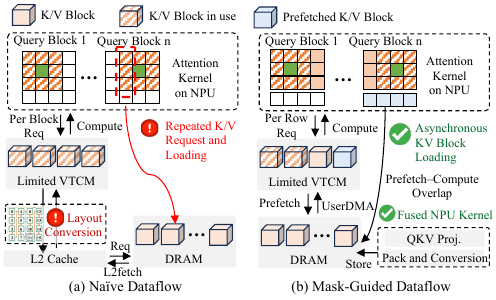}
    \caption{(a) Naive Dataflow with repeated K/V loading and layout conversion; (b) Mask-Guided Dataflow with K/V reuse, prefetching, and fused QKV packing.}
    \label{fig:dataflow}
\end{figure}

\subsection{Mask-Guided Dataflow Optimization}

Heterogeneous XPU collaboration in Section~\ref{sec:4.3} limits anchor-step calibration overhead, but NPU attention must still move Q, K, and V from DRAM into VTCM and arrange them in HMX's tile layout. The full layer's K/V working set cannot reside on-chip. Loading retained blocks independently therefore repeats transfers of overlapping image-halo and text K/V data. Figure~\ref{fig:dataflow}(a) depicts these per-block requests and the separate layout conversion, both of which can leave HMX waiting for data.

Table~\ref{tab:kv_movement} shows that with independent per-query loading, data wait and K/V repacking take 4.52~ms and 5.45~ms, respectively. Cross-query reuse reduces these costs to 2.11~ms and 3.40~ms, but both remain significant. \ours therefore uses the predictable access pattern of the static mask to group and prefetch K/V blocks, while producing Q/K/V in the layout required by attention.

\paragraph{Asynchronous KV Block Loading.}
The static halo mask in Section~\ref{sec:4.2} determines each query block's K/V access set before inference, but transferring blocks individually still creates many small requests. \ours groups query blocks in the same grid row into a \emph{query row}, gathers their required K/V blocks, and merges contiguous addresses into transfer intervals. Using tensor strides, block width, and target VTCM addresses, it constructs two-dimensional UserDMA requests that move multiple token rows with one submission. Figure~\ref{fig:dataflow}(b) shows these per-row requests in place of the per-block requests.

Merged requests alone cannot prevent adjacent query rows from loading the same blocks again. \ours retains image K/V needed by later rows and keeps shared text K/V resident when VTCM capacity permits. It loads the remaining image K/V by row or small batch, reusing available blocks while keeping the on-chip working set within capacity.

To overlap transfers with computation, \ours uses two buffers for transfers and two for the packed K/V cache. While HMX processes the current query row, UserDMA prefetches K/V missing from the next row into the other buffer. The runtime waits for that transfer before consuming the prefetched blocks, then alternates between buffers. As illustrated in Figure~\ref{fig:dataflow}(b), this schedule preserves blocks in use while loading the next row alongside NPU computation.

\paragraph{Fused NPU Packing and Conversion Kernel.}
Double buffering hides transfer time, but HMX still requires K/V in a tile-first layout. With unpacked inputs, the system must pack loaded data and cache it for neighboring query rows. If QKV projection first writes row-major tensors and the CPU then applies normalization or RoPE, separate rearrangement adds memory accesses and cross-processor synchronization.

\ours instead moves packing to the QKV projection's write-out stage and fuses projection with its post-processing on the NPU. The operator converts FP32 activations to FP16, performs matrix multiplication and bias addition, applies Q/K RMSNorm and the required RoPE, and writes packed Q, K, and V. It processes image and text inputs separately, placing them in a unified layout at 32-token tile boundaries without an intermediate row-major concatenation. Figure~\ref{fig:dataflow}(b) shows packing and conversion alongside QKV projection, replacing the separate conversion stage in (a). Subsequent DMA requests transfer packed data, which the attention kernel accesses through tile offsets without further packing.

Compared with block reuse alone, \ours reduces DRAM reads from 216.4~MB to 51.36~MB and transfer stalls from 2.11~ms to 0.02~ms while eliminating rearrangement before attention. Compute time remains largely unchanged; the gain comes primarily from lower data preparation costs.

%% file: _txt/5_implementation.tex
\section{\textbf{Implementation}}
\label{sec:impl}

{\ours is implemented on the ggml~\cite{ggml} compute graph of \texttt{stable-diffusion.cpp}~\cite{sdcpp}, with HTP kernels compiled using the LLVM~\cite{llvm} toolchain in Hexagon SDK 6.0.0.2~\cite{hexagonsdk}. Our implementation consists of three modules: Cross-XPU Execution, Dataflow Optimization, and Error Compensation. The CPU executes and schedules the graph, the HTP runs operators built around HMX and VTCM~\cite{hexagonnpu} through the open-source \texttt{htp-ops-lib}~\cite{htpopslib}, and OpenCL~\cite{opencl} runs sparse attention asynchronously. Keeping the graph in ggml instead of QNN~\cite{qnn} allows the mask format, block size, and sparse plan to change across denoising steps.\par}

{\textbf{\textit{Cross-XPU Execution.}} The CPU, CDSP, and OpenCL device operate on shared buffers. We allocate HTP buffers with \codeid{rpcmem_alloc}~\cite{dmabuf}, obtain their descriptors with \codeid{rpcmem_to_fd}, and map them into the CDSP address space with \codeid{fastrpc_mmap}~\cite{fastrpc}. Tensor addresses are passed as descriptor--offset pairs. The OpenCL side imports the same descriptors as \codeid{cl_mem} using \codeid{CL_MEM_DMABUF_HOST_PTR_QCOM} and \codeid{CL_MEM_USE_HOST_PTR}~\cite{clqcomionhostptr}, avoiding copies between processors. Imported objects are cached for buffers reused across graphs. We track mapping lifetimes and backing-store references separately, and unmap a buffer only after its last graph alias is released.\par}

\textbf{\textit{Dataflow Optimization.}} We instantiate a fixed block-sparse mask from the token layout and static-halo policy, and store its block size, mask format, and halo parameters as attention-node attributes for reuse across denoising steps. On the HTP, dynamic KV sparse attention constructs 2D UserDMA descriptors for the retained K/V blocks and transfers them from shared DRAM to VTCM. Its transfer plan follows each query block's mask, while two staging banks prefetch the next K/V group during the current matrix computation. A second operator fuses QKV projection, bias, and Q/K normalization, applies rotary embedding on branches that use it, and writes Q, K, and V directly in the packed layout expected by attention. This removes the intermediate tensors and row-major round trip.

\textbf{\textit{Error Compensation.}} At each dense residual anchor, the attention node evaluates dense and sparse branches. Exact correction computes $\rho$ from their LSEs. Otherwise, we use the preset scaling factor. We cache $\rho$ and the difference between the dense output and corrected sparse output once per anchor, then apply the cached correction to subsequent sparse steps without rerunning dense attention. To overlap the anchor branches, the graph thread submits sparse attention to one ordered OpenCL queue, flushes it, publishes its completion event, and starts dense attention on the HTP. A second queue submits the $\rho$ and residual kernels with the sparse event as a dependency. The shared Q/K/V range remains leased until sparse attention completes. Sparse O/LSE staging buffers and the dense-output lease remain held until the residual completes.

%% file: _txt/6_experiment.tex
\section{\textbf{Experiments}}
\label{sec:experiments}

\begin{figure*}[t]
  \centering
  \includegraphics[width=\textwidth]{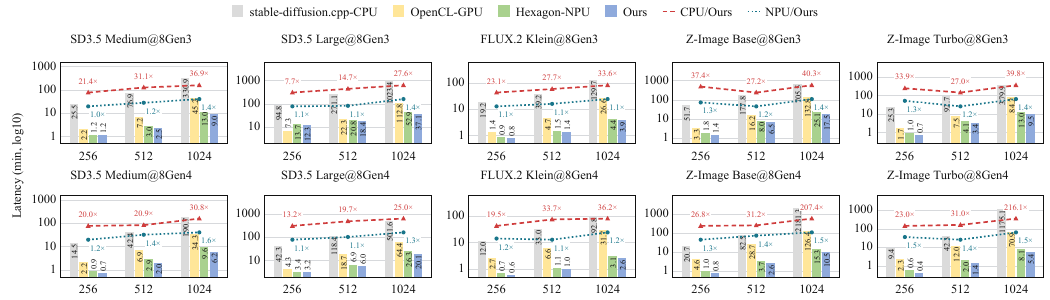}
  \caption{End-to-end denoising latency of \ours{} and baseline backends across DiT models and mobile SoCs.}
  \label{fig:end-to-end-latency}
\end{figure*}

\begin{figure*}[t]
  \centering
  \includegraphics[width=\textwidth]{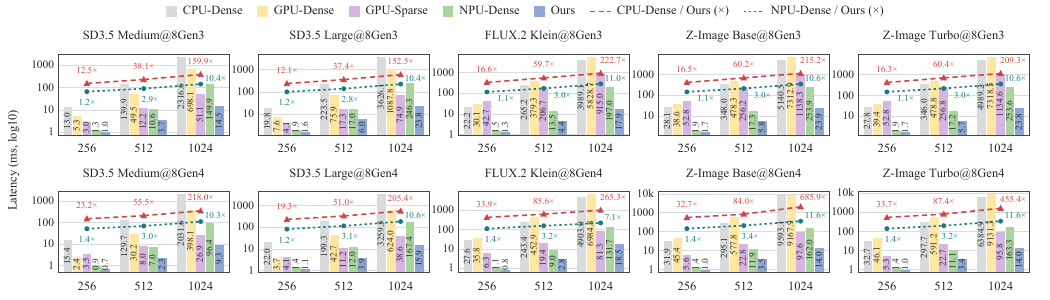}
  \caption{Attention kernel latency of \ours{} and the baseline backends across DiT models and mobile SoCs.}
  \label{fig:attention-kernel-8gen3}
\end{figure*}

\subsection{Experiment Setup}
\label{sec:experiment-setup}

\paragraph{Hardware platforms.}
We evaluate \ours on two mobile devices equipped with different generations of Snapdragon SoCs. All devices run Android 13. The first device is a OnePlus Ace 5 with a Snapdragon 8 Gen 3 SoC, 24~GB of memory, and 1~TB of UFS~4.0 storage. The second is a OnePlus 13 with a Snapdragon 8 Gen 4 SoC, 24~GB of memory, and 1~TB of UFS~4.0 storage. These platforms cover multiple generations of Qualcomm mobile hardware and let us assess the portability of \ours across heterogeneous devices.

\paragraph{Models and resolutions.}
We select representative text-to-image DiT models with different parameter scales, network architectures, and denoising schedules. The MMDiT family includes Stable Diffusion 3.5 Medium ~\cite{sd3}, Stable Diffusion 3.5 Large, and FLUX.2~[klein]~\cite{flux2}. The S3-DiT family includes Z-Image~\cite{zimage} and its distilled variant, Z-Image-Turbo. All models are evaluated at $256\times256$, $512\times512$, and $1024\times1024$ output resolutions. Unless otherwise stated, each model uses its standard sampling schedule and precision on all backends.Table~\ref{tab:model_config} summarizes the architecture and the sampling configuration of the five models.

\paragraph{Datasets and metrics.}
To quantify the effect of sparse inference on generation quality, we use several established text-to-image benchmarks. GenEval~\cite{geneval} measures compositional generation ability. HPSv2~\cite{hpsv2} and ImageReward~\cite{imagereward} measure alignment with human preferences. On MJHQ-30K~\cite{mjhq30k}, we report FID~\cite{fid} and CLIP Score~\cite{clipscore} to evaluate the distributional quality of generated images and text--image consistency, respectively. All generation-quality evaluations are conducted on four NVIDIA L40 GPUs.

\paragraph{Comparison baselines.}
We compare \ours with baselines at two levels. For end-to-end inference, we use the CPU, OpenCL-GPU, and Hexagon-NPU backends of Stable-diffusion.cpp~\cite{sdcpp}, representing execution on the mobile CPU, GPU, and NPU, respectively. For sparse attention, we compare with PISA~\cite{pisa} at 80\% sparsity and SpargeAttn~\cite{spargeattn} at 40\% sparsity, following their evaluated configurations.

\paragraph{Settings.}
We use dense attention for the first 10\% of denoising steps as a warm-up, then switch to sparse attention with dense anchors at five-step intervals. Sparse attention uses $32\times32$ query--key tiles and a fixed $3\times3$ image halo ($r=1$), while all connections involving text tokens remain dense. All reported latencies are averaged over five runs.

\input{_tab/tab_model_config}

\subsection{Latency Evaluation}
\label{sec:latency-evaluation}
\subsubsection{End-to-End Denoising Latency}
\label{sec:end-to-end}
\label{sec:denoising-latency}

We compare \ours{} with the CPU, OpenCL-GPU, and Hexagon-NPU backends of SDCpp. As shown in Figure~\ref{fig:end-to-end-latency}, \ours{} outperforms Hexagon-NPU in all 30 model--resolution--device settings and achieves the lowest latency in 29 of them. At $1024\times1024$, it delivers a $1.15$--$1.45\times$ speedup over Hexagon-NPU on the 8 Gen 3 device and a $1.21$--$1.56\times$ speedup on the 8 Gen 4 device. The gains are most pronounced at the highest resolution, where the larger attention workload provides more opportunity for sparse execution to reduce denoising time.

\subsubsection{End-to-End Latency of Sparse-Attention Strategies}
\label{sec:sparse-stage-overhead}

We compare the end-to-end denoising latency of \ours{} with dense attention, PISA~\cite{pisa}, SpargeAttn~\cite{spargeattn}, and a static sparse-attention baseline across the five models. Table~\ref{tab:speed_comparison} shows that \ours{} achieves an average $1.40\times$ speedup over dense attention, exceeding the speedups of PISA and SpargeAttn. This result suggests that a fixed, hardware-aligned sparsity pattern and amortized anchor calibration translate attention savings into end-to-end gains more effectively than per-step sparse selection.

\input{_tab/tab_sparse_route}

\subsection{Kernel Performance Evaluation}
\label{sec:kernel-performance}

\subsubsection{Attention Kernel}
\label{sec:attention-kernel}

We benchmark a single attention kernel on two devices over five models at different output image resolutions, with all implementations sharing the same Q/K/V tensors, data precision, block size and tensor layout. Figure~\ref{fig:attention-kernel-8gen3} shows that \ours{} is the fastest implementation in every measured configuration on both devices, and that its margin over the best alternative grows with the resolution, reaching $2.50$--$3.27\times$ on the 8 Gen 3 and $2.68$--$3.38\times$ on the 8 Gen 4 device at $512\times512$ and up to $11.5\times$ and $7.0\times$ at $1024\times1024$.

\subsubsection{Energy Efficiency}
\label{sec:energy-efficiency}

We measure the energy of a single attention kernel on the five models at $1024\times1024$ resolution. Table~\ref{tab:single-kernel-energy} shows that \ours{} consumes $0.72$--$1.18$~J, an order of magnitude below CPU-dense and $2$--$7\times$ below GPU-dense, and removing the dense anchor correction lowers it to $0.05$--$0.09$~J, the lowest of all implementations.

\input{_tab/tab_energy}

\subsection{Generation Quality}
\label{sec:generation-quality}

We evaluate generation quality at $1024\times1024$ across five models, comparing \ours{} with dense attention and sparse-attention baselines. As shown in Table~\ref{tab:quality_compact}, \ours{} closely matches dense attention on GenEval: it scores $0.35$--$1.05$ points higher on four models and only $0.18$ points lower on SD3.5 Medium. It also achieves the highest ImageReward among sparse methods on SD3.5 Large and FLUX.2 [klein], with scores of $0.8878$ and $0.9243$, respectively. Figure~\ref{fig:qualitative} provides representative visual comparisons. Together, these results suggest that \ours{} largely preserves generation quality while accelerating inference through sparse attention.

% Rotated row label (kept here so this file compiles on its own).
% It centres the label on the row of images and reports zero height/depth so
% that it does not grow the row.  0.0775\columnwidth = half of 0.155\columnwidth,
% the width given to every image below.
\providecommand{\qualrowlabel}[1]{%
  \raisebox{\dimexpr 0.0775\columnwidth - 0.5\height + 0.5\depth\relax}[0pt][0pt]{%
    \scriptsize\rotatebox[origin=c]{90}{#1}}%
}
\begin{figure}[t]
  \centering
  \setlength{\tabcolsep}{2pt}
  \renewcommand{\arraystretch}{1.15}
  \begin{tabular}{@{}l@{\hspace{3pt}}ccccc@{}}
    & {\scriptsize Dense} & {\scriptsize PISA} & {\scriptsize Sparge} & {\scriptsize Static} & {\scriptsize \textbf{Ours}} \\
    \qualrowlabel{SD3.5-M}
      & \includegraphics[width=0.155\columnwidth]{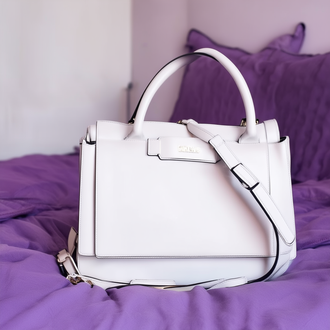} &
      \includegraphics[width=0.155\columnwidth]{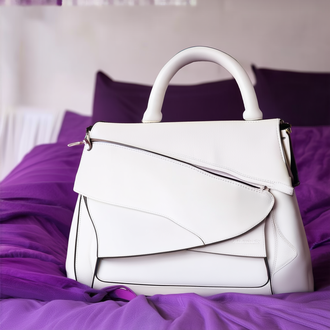} &
      \includegraphics[width=0.155\columnwidth]{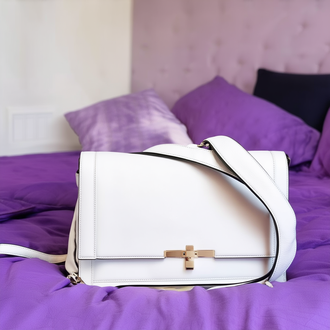} &
      \includegraphics[width=0.155\columnwidth]{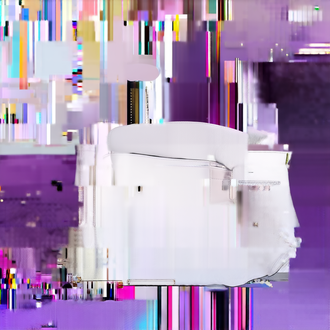} &
      \includegraphics[width=0.155\columnwidth]{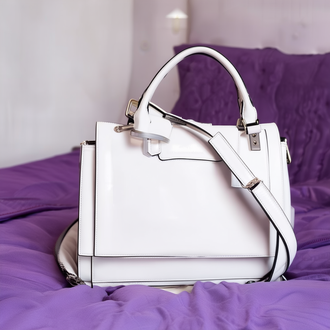} \\
    \qualrowlabel{SD3.5-L}
      & \includegraphics[width=0.155\columnwidth]{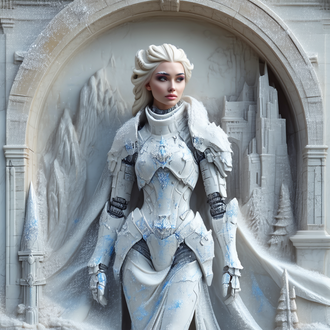} &
      \includegraphics[width=0.155\columnwidth]{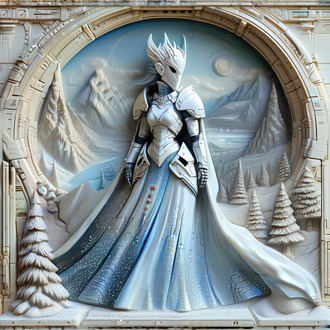} &
      \includegraphics[width=0.155\columnwidth]{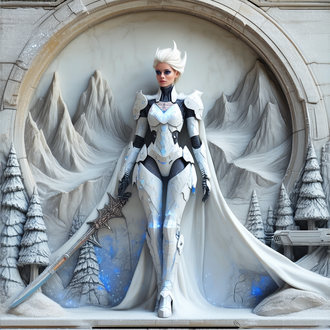} &
      \includegraphics[width=0.155\columnwidth]{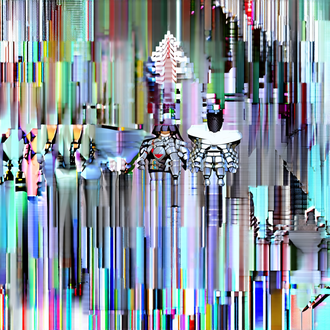} &
      \includegraphics[width=0.155\columnwidth]{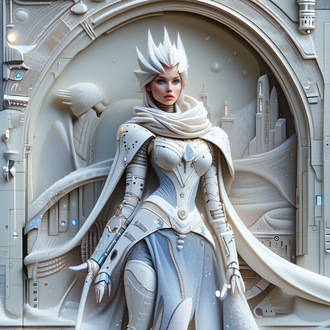} \\
    \qualrowlabel{Z-Image}
      & \includegraphics[width=0.155\columnwidth]{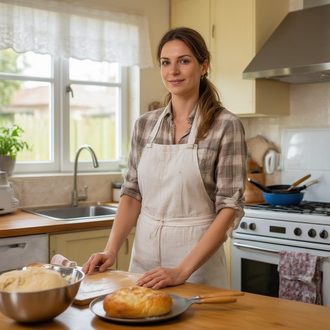} &
      \includegraphics[width=0.155\columnwidth]{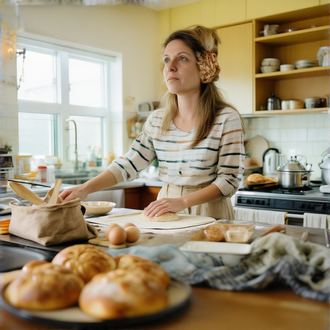} &
      \includegraphics[width=0.155\columnwidth]{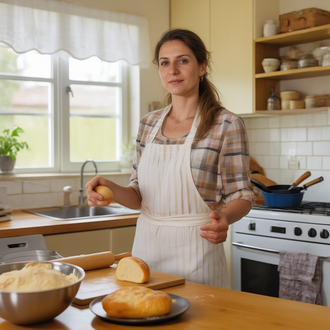} &
      \includegraphics[width=0.155\columnwidth]{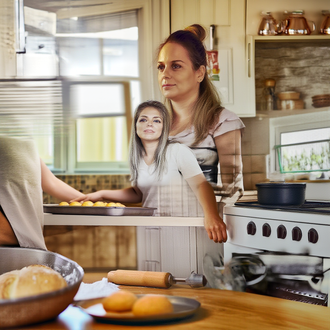} &
      \includegraphics[width=0.155\columnwidth]{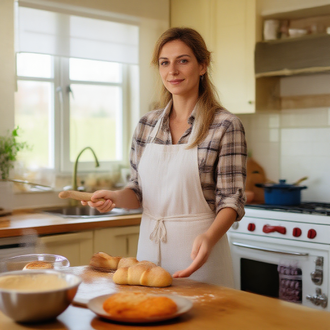} \\
  \end{tabular}
  \caption{Qualitative comparison of \ours{} and the baselines on three prompts, one per row: SD3.5-Medium, SD3.5-Large and Z-Image.}
  \label{fig:qualitative}
\end{figure}

\input{_tab/tab_quality}

\subsection{Ablation Study}
\label{sec:ablation}

\subsubsection{Component Ablation}
\label{sec:component-ablation}

\begin{figure}[t]
  \centering
  \includegraphics[width=\columnwidth]{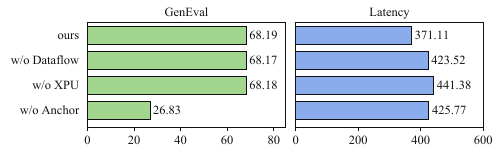}
  \caption{Component ablation. Each build adds one technique to the previous one.}
  \label{fig:component-ablation}
\end{figure}

We enable the proposed techniques one at a time on SD3.5 Medium at $1024\times1024$. Figure~\ref{fig:component-ablation} reports the GenEval score and the end-to-end latency of the five versions: the complete \ours{} reduces the latency from $578.6$~s to $371.1$~s, a $1.56\times$ speedup, with GenEval essentially unchanged at $68.19\%$ against $68.37\%$, whereas removing the anchor compensation collapses GenEval to $26.83\%$.

\subsubsection{Anchor Step Interval}
\label{sec:anchor-step}

We vary only the anchor interval $p$, which inserts one anchor step every $p$ sparse steps, on SD3.5 Medium at $1024\times1024$, and report, for $p=2$ to $p=6$, GenEval, HPSv2.1, ImageReward and the end-to-end latency on the 8 Gen 4 device in Table~\ref{tab:anchor-step}. A sparser anchoring trades a small quality loss for a modest latency gain: from $p=2$ to $p=5$ the latency falls from $412.32$~s to $371.11$~s while GenEval moves only from $68.36\%$ to $68.19\%$.

\input{_tab/tab_abaltion_p}

\subsubsection{Heterogeneous Strategy}
\label{sec:hetero-ablation}

\begin{figure*}[t]
  \centering
  \includegraphics[width=\textwidth]{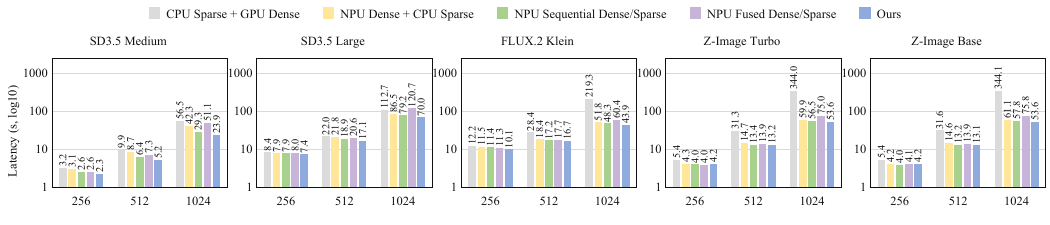}
  \caption{Heterogeneous-strategy ablation for a single denoising step on the Snapdragon 8 Gen 4 device. }
  \label{fig:hetero-ablation}
\end{figure*}

We fix one denoising step on the Snapdragon 8 Gen 4 device and compare four alternative execution strategies against \ours: \emph{npu divided} computes the attention twice, separately, on the NPU; \emph{npu fused} executes a single fused dense-and-sparse attention on the NPU; \emph{cpu + gpu} executes the sparse attention on the CPU and the dense attention on the GPU; and \emph{npu + cpu} executes the dense attention on the NPU and the sparse attention on the CPU. Figure~\ref{fig:hetero-ablation} shows that \ours{} is the fastest strategy in $13$ of the $15$ configurations: at $1024\times1024$ it is $1.05$--$1.23\times$ faster than the best of the four alternatives.

\begin{figure*}[t]
  \centering
  \includegraphics[width=\textwidth]{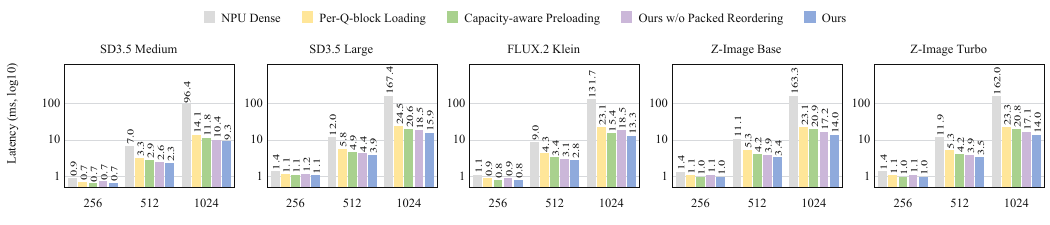}
  \caption{KV load ablation of the attention kernel on the Snapdragon 8 Gen 4 device. }
  \label{fig:kv-ablation}
\end{figure*}

\subsubsection{KV Load/Unload}
\label{sec:kv-ablation}

We compare four NPU sparse-attention loading schemes, including \emph{chunk} and \emph{baseline}: \emph{chunk} computes one Q block at a time and then loads the corresponding KV blocks; \emph{baseline} first attempts to load all KV blocks and loads them if the VTCM capacity is sufficient, otherwise attempts to load all K blocks and loads them if the VTCM capacity is sufficient, and falls back to \emph{chunk} otherwise; and \emph{no packed} refers to our method without the repacking optimization proposed in \S 4.4.2. Figure~\ref{fig:kv-ablation} shows that \ours{} reduces the kernel latency against dense loading by $9.9$--$12.3\times$ at $1024\times1024$, with a median of $3.25\times$ over the $27$ configurations, and that in the median it is $1.44\times$ faster than chunk and $1.18\times$ faster than baseline.

\subsubsection{Performance at Different Resolutions}
\label{sec:resolution-ablation}

We measure the end-to-end latency of the SD3.5-medium model at different resolutions on the 8 Gen 4 device. We cover two pairs of resolutions that contain the same number of pixels but differ in spatial arrangement, together with a high-resolution case at $2048\times2048$. Table~\ref{tab:resolution-ablation} reports the end-to-end latency of \ours{}: it is $1.99\times$ and $1.94\times$ faster than dense attention on the first pair, $1.52\times$ and $1.56\times$ on the second.

\input{_tab/tab_resolution}

\subsection{Orthogonality with Other Acceleration Methods}
\label{sec:orthogonality}

To verify orthogonality with other acceleration methods, we combine \ours{} with TeaCache~\cite{teacache}, DBCache~\cite{cachedit} and TaylorSeer~\cite{taylorseer}. As Table~\ref{tab:caching_speed_comparison} shows, adding TaylorSeer lowers the GenEval score only slightly, to $0.6423$, while the speedup on the 8 Gen 4 reaches $1.74\times$. The results show that \ours{} is compatible with existing optimization methods.

\input{_tab/tab_cache}

%% file: _tab/tab_model_config.tex
% Requires: \usepackage{booktabs}
\begin{table}[t]
\centering
\caption{Architecture and sampling configuration of the evaluated DiT models.}
\label{tab:model_config}
\scriptsize
\small
\setlength{\tabcolsep}{2.5pt}
\renewcommand{\arraystretch}{1.0}

\resizebox{\columnwidth}{!}{
\begin{tabular}{@{}lcccccc@{}}
\toprule
\textbf{Model} & \textbf{Architecture} & \textbf{Params} &
\textbf{Heads} & \textbf{Head Dim} & \textbf{Layers} & \textbf{Steps} \\
\midrule
SD3.5 Medium    & MMDiT-X        & 2.5B & 24 & 64  & 24 & 20 \\
SD3.5 Large     & MMDiT          & 8.1B & 38 & 64  & 38 & 20 \\
Z-Image         & S3-DiT         & 6B   & 30 & 128 & 32 & 15 \\
Z-Image Turbo   & S3-DiT         & 6B   & 30 & 128 & 32 & 8  \\
FLUX.2 [klein]  & RF Transformer & 4B   & 24 & 128 & 25 & 4  \\
\bottomrule
\end{tabular}
}
\end{table}

%% file: _tab/tab_sparse_route.tex
% Requires: \usepackage{booktabs}
\begin{table}[t]
\centering
\caption{End-to-end denoising latency (s) of dense attention and three sparse-attention
strategies at $1024\times1024$ resolution on the Snapdragon 8 Gen 4. }
\label{tab:speed_comparison}
\small
\setlength{\tabcolsep}{2pt}
\renewcommand{\arraystretch}{1.05}

\begin{tabular}{@{}l|ccccc|c@{}}
\toprule
\textbf{Latency (s) $\downarrow$}
& \textbf{Dense}
& \textbf{PISA}
& \textbf{Sparge}
& \textbf{Static}
& \textbf{Ours}
& \textbf{Speedup} \\
\midrule
SD3.5 Medium    & 579  & 1446 & 876  & 325  & 371  & $1.56\times$ \\
SD3.5 Large     & 1579 & 3183 & 2185 & 1385 & 1206 & $1.31\times$ \\
Z-Image         & 919  & 3030 & 1144 & 536  & 631  & $1.46\times$ \\
Z-Image-Turbo   & 483  & 1239 & 616  & 292  & 326  & $1.48\times$ \\
FLUX.2-klein    & 186  & 433  & 295  & 147  & 154  & $1.21\times$ \\
\midrule
\textbf{Mean Speedup}
& $1.00\times$ & $0.40\times$ & $0.72\times$ & $1.51\times$ & $1.40\times$ & \\
\bottomrule
\end{tabular}
\end{table}

%% file: _tab/tab_energy.tex
% Requires: \usepackage{booktabs}
\begin{table}[t]
\centering
\caption{Single-kernel energy consumption across DiT models and execution configurations.}
\label{tab:single-kernel-energy}
\small
\setlength{\tabcolsep}{3pt}
\renewcommand{\arraystretch}{1.05}

\begin{tabular}{@{}l|cccccc@{}}
\toprule
Model
& \shortstack{CPU\\Dense}
& \shortstack{GPU\\Dense}
& \shortstack{GPU\\Sparse}
& \shortstack{NPU\\Dense}
& \shortstack{GPU/NPU\\Anchor}
& \shortstack{NPU\\Sparse} \\
\midrule
SD3.5 Medium
& 18.32 & 2.35 & 0.48 & 0.45 & 0.76 & 0.05 \\

SD3.5 Large
& 12.88 & 3.77 & 0.77 & 0.79 & 1.18 & 0.08 \\

FLUX.2-klein
& 12.17 & 4.20 & 0.23 & 0.53 & 0.72 & 0.07 \\

Z-Image
& 15.16 & 5.37 & 0.28 & 0.73 & 0.90 & 0.09 \\

Z-Image-Turbo
& 14.98 & 5.32 & 0.27 & 0.71 & 0.85 & 0.08 \\
\bottomrule
\end{tabular}
\end{table}

%% file: _tab/tab_quality.tex
% Requires: \usepackage{booktabs}
\begin{table}[t]
\centering
\caption{Generation quality comparison across attention strategies.
FID and CLIP are averaged over Landscape, People, and Food.
$\Delta$ denotes the absolute change relative to Dense.
For $\uparrow$ metrics, positive values indicate improvement; for FID$\downarrow$, negative values indicate improvement.}
\label{tab:quality_compact}
\small
\setlength{\tabcolsep}{2pt}
\renewcommand{\arraystretch}{1.05}

\begin{tabular}{@{}l|ccccc|c@{}}
\toprule
\textbf{GenEval $\uparrow$}
& \textbf{Dense}
& \textbf{PISA}
& \textbf{Sparge}
& \textbf{Static}
& \textbf{Ours}
& \textbf{$\Delta$} \\
\midrule
SD3.5 Medium  & 68.37 & 64.55 & 68.19 & 26.83 & 68.19 & \textbf{-0.18} \\
SD3.5 Large   & 71.31 & 70.06 & 70.54 & 23.66 & 71.88 & \textbf{+0.57} \\
Z-Image       & 75.35 & 73.30 & 75.13 & 65.88 & 76.26 & \textbf{+0.91} \\
Z-Image-Turbo & 75.49 & 72.49 & 75.47 & 63.14 & 76.54 & \textbf{+1.05} \\
FLUX.2-klein  & 82.66 & 84.87 & 82.40 & 70.60 & 83.01 & \textbf{+0.35} \\
\midrule
\textbf{Avg. $\Delta$}
& -- & \textbf{-1.58} & \textbf{-0.29} & \textbf{-24.61}
& \textbf{+0.54} & \\
\bottomrule

\addlinespace[2pt]

\toprule
\textbf{HPSv2.1 $\uparrow$}
& \textbf{Dense}
& \textbf{PISA}
& \textbf{Sparge}
& \textbf{Static}
& \textbf{Ours}
& \textbf{$\Delta$} \\
\midrule
SD3.5 Medium  & 29.18 & 27.60 & 29.35 & 20.46 & 29.53 & \textbf{+0.35} \\
SD3.5 Large   & 29.91 & 28.51 & 29.81 & 19.35 & 30.12 & \textbf{+0.21} \\
Z-Image       & 27.74 & 26.94 & 27.75 & 25.50 & 27.72 & \textbf{-0.02} \\
Z-Image-Turbo & 29.83 & 30.01 & 29.86 & 28.59 & 29.86 & \textbf{+0.03} \\
FLUX.2-klein  & 30.09 & 29.81 & 30.10 & 28.39 & 29.21 & \textbf{-0.88} \\
\midrule
\textbf{Avg. $\Delta$}
& -- & \textbf{-0.78} & \textbf{+0.02} & \textbf{-4.89}
& \textbf{-0.06} & \\
\bottomrule

\addlinespace[2pt]

\toprule
\textbf{ImageReward $\uparrow$}
& \textbf{Dense}
& \textbf{PISA}
& \textbf{Sparge}
& \textbf{Static}
& \textbf{Ours}
& \textbf{$\Delta$} \\
\midrule
SD3.5 Medium  & 0.88 & 0.86 & 0.85 & -0.75 & 0.86 & \textbf{-0.02} \\
SD3.5 Large   & 0.88 & 0.69 & 0.88 & -1.32 & 0.89 & \textbf{+0.01} \\
Z-Image       & 0.69 & 0.60 & 0.66 & 0.24 & 0.66 & \textbf{-0.03} \\
Z-Image-Turbo & 0.70 & 0.77 & 0.69 & 0.47 & 0.69 & \textbf{-0.01} \\
FLUX.2-klein  & 0.91 & 0.91 & 0.92 & 0.55 & 0.92 & \textbf{+0.01} \\
\midrule
\textbf{Avg. $\Delta$}
& -- & \textbf{-0.05} & \textbf{-0.01} & \textbf{-0.97}
& \textbf{-0.01} & \\
\bottomrule

\addlinespace[2pt]

\toprule
\textbf{FID $\downarrow$}
& \textbf{Dense}
& \textbf{PISA}
& \textbf{Sparge}
& \textbf{Static}
& \textbf{Ours}
& \textbf{$\Delta$} \\
\midrule
SD3.5 Medium  & 118.70 & 104.92 & 150.24 & 139.90 & 107.60 & \textbf{-11.10} \\
SD3.5 Large   & 119.07 & 102.11 & 142.20 & 190.74 & 108.07 & \textbf{-11.00} \\
Z-Image       & 121.23 & 105.55 & 105.38 & 125.54 & 104.97 & \textbf{-16.26} \\
Z-Image-Turbo & 127.11 & 111.00 & 112.17 & 129.37 & 114.59 & \textbf{-12.53} \\
FLUX.2-klein  & 122.52 & 116.90 & 118.37 & 125.42 & 114.90 & \textbf{-7.61} \\
\midrule
\textbf{Avg. $\Delta$}
& -- & \textbf{-13.63} & \textbf{+3.95} & \textbf{+20.47}
& \textbf{-11.70} & \\
\bottomrule

\addlinespace[2pt]

\toprule
\textbf{CLIP $\uparrow$}
& \textbf{Dense}
& \textbf{PISA}
& \textbf{Sparge}
& \textbf{Static}
& \textbf{Ours}
& \textbf{$\Delta$} \\
\midrule
SD3.5 Medium  & 81.17 & 80.07 & 81.33 & 68.84 & 81.28 & \textbf{+0.11} \\
SD3.5 Large   & 81.45 & 80.84 & 81.67 & 65.50 & 81.65 & \textbf{+0.20} \\
Z-Image       & 80.56 & 80.25 & 80.52 & 78.48 & 80.81 & \textbf{+0.25} \\
Z-Image-Turbo & 79.35 & 79.98 & 79.38 & 79.17 & 80.25 & \textbf{+0.90} \\
FLUX.2-klein  & 79.67 & 80.13 & 79.72 & 78.82 & 83.72 & \textbf{+4.05} \\
\midrule
\textbf{Avg. $\Delta$}
& -- & \textbf{-0.19} & \textbf{+0.08} & \textbf{-6.28}
& \textbf{+1.10} & \\
\bottomrule
\end{tabular}
\end{table}

%% file: _tab/tab_abaltion_p.tex
% Requires: \usepackage{booktabs}
\begin{table}[t]
\centering
\caption{Generation quality and latency of \ours{} under different anchor step intervals $p$ on SD3.5 Medium at $1024\times1024$ resolution and 20 steps.}
\label{tab:anchor-step}
\small
\setlength{\tabcolsep}{6pt}
\renewcommand{\arraystretch}{1.05}

\begin{tabular}{@{}c|cccc@{}}
\toprule
$p$
& GenEval $\uparrow$
& HPSv2.1 $\uparrow$
& ImageReward $\uparrow$
& Latency (s) $\downarrow$ \\
\midrule
2 & 68.36 & 29.63 & 0.8813 & 412.32 \\
3 & 68.31 & 29.60 & 0.8725 & 394.88 \\
4 & 68.24 & 29.55 & 0.8604 & 386.40 \\
5 & 68.19 & 29.53 & 0.8575 & 371.11 \\
6 & 66.07 & 29.42 & 0.8219 & 367.09 \\
\bottomrule
\end{tabular}
\end{table}

%% file: _tab/tab_resolution.tex
% Requires: \usepackage{booktabs}
\begin{table}[t]
  \centering
  \caption{Latency of SD3.5 Medium at different resolutions.}
  \label{tab:resolution-ablation}
  \small
  \setlength{\tabcolsep}{6pt}
  \begin{tabular}{@{}l c c c c@{}}
    \toprule
     & $1920\times1080$ & $1080\times1920$ & $896\times1152$ & $1152\times896$ \\
    \midrule
    Dense   & 1565.88 & 1556.36 & 493.32 & 495.24 \\
    \ours{} & 786.14  & 803.97  & 324.48 & 317.02 \\
    \bottomrule
  \end{tabular}
\end{table}

%% file: _tab/tab_cache.tex
% Requires: \usepackage{booktabs}
\begin{table}[t]
\centering
\caption{GenEval scores, inference latency and speedup of orthogonal caching
on the DiT.xPU backend for the SD3.5-medium model at $1024\times1024$ resolution.}
\label{tab:caching_speed_comparison}
\small
\setlength{\tabcolsep}{10pt}
\renewcommand{\arraystretch}{1.05}

\begin{tabular}{@{}l|ccc@{}}
\toprule
\textbf{Method} & \textbf{GenEval $\uparrow$} & \textbf{Latency (s) $\downarrow$} & \textbf{Speedup $\uparrow$} \\
\midrule
Dense                & 0.6837 & 579    & 1.00$\times$ \\
ours             & 0.6819 & 371.11 & 1.56$\times$ \\
w/ TeaCache   & 0.6451 & 361.57 & \textbf{1.60$\times$} \\
w/ DBCache    & 0.6399 & 333.89 & \textbf{1.73$\times$} \\
w/ TaylorSeer & 0.6423 & 333.14 & \textbf{1.74$\times$} \\
\bottomrule
\end{tabular}
\end{table}

%% file: _txt/7_related_work.tex
\section{\textbf{Related Work}}
\label{sec:related work}

\subsection{Inference Optimization on Mobile SoCs}
\label{sec:rw-mobile-npu}

On-device inference has received growing attention. Mobile SoCs integrate CPUs, GPUs, and NPUs, and existing work primarily accelerates LLM inference by distributing computation across these processors~\cite{heterollm,llmnpu,powerinfer2}. For example, HeteroLLM partitions computation between the GPU and NPU to reduce data copies, but still runs most attention operations on the GPU. llm.npu assigns quantized linear layers to the NPU while leaving attention on the CPU or GPU. PowerInfer-2 also uses the NPU for dense attention, but does not reduce the amount of attention computation. More closely related to our work, shadowAttn~\cite{shadowattn} uses the NPU to identify important tokens and then performs sparse attention on the CPU or GPU. It updates token selection as the input changes, while the sparse computation itself remains outside the NPU. For DiTs~\cite{dit}, which process the full visual token grid at every denoising step, existing engines such as stable-diffusion.cpp~\cite{sdcpp} support on-device inference but do not target block-sparse attention on mobile NPUs. \ours maps block-sparse attention to the NPU's fixed compute granularity to improve execution efficiency.

\subsection{Sparse Inference Optimization for DiTs}
\label{sec:rw-sparse-attention}

Prior work has reduced the cost of DiT inference through techniques such as caching~\cite{teacache,cachedit,taylorseer} and distillation~\cite{distill}. Sparse attention directly reduces the attention computation at each denoising step by omitting selected token interactions~\cite{spargeattn,sparsevideogen,radialattn}. These methods can be grouped by whether the retained interactions are updated at runtime.
    
    \textbf{Static sparse attention} determines which interactions to retain before inference, using criteria such as spatial proximity or similarity across denoising steps~\cite{diffastattn}. The resulting computation and memory accesses are predictable, enabling regular execution. However, important interactions vary with the input and denoising state, so a fixed pattern may omit them as sparsity increases.
    
    \textbf{Dynamic sparse attention} such as PISA~\cite{pisa} and SpargeAttn estimates the importance of attention blocks at runtime and selects which interactions to compute. Others adapt the sparsity pattern to differences across heads and layers~\cite{diffastattnv2}. This flexibility introduces overhead from pattern selection, data rearrangement, and irregular memory accesses. Existing implementations rely primarily on custom GPU kernels to manage these costs and are difficult to map directly onto mobile NPUs with fixed compute granularity and limited on-chip memory. Trainable variants instead learn or fine-tune a fixed pattern before inference~\cite{spargeattn2,vsa,sla}, and permutation-based methods reorder the tokens so that the retained blocks become contiguous~\cite{sparsevgen2}. Unlike prior methods, ours aligns static block sparsity with NPU matrix tiles and calibrates sparse outputs using dense attention at selected anchor steps.

%% file: _txt/8_conclusion.tex
\section{\textbf{Conclusion}}
\label{sec:conclusion}

We presented DiT.XPU, a sparse inference system for diffusion transformers on mobile heterogeneous SoCs. DiT.XPU combines an NPU-aligned halo mask with periodic dense anchors to compensate for omitted attention contributions, while cross-XPU execution and mask-guided K/V dataflow limit the cost of calibration and data movement. Across five models, three resolutions, and two mobile SoCs, it reduces generation latency relative to the NPU baseline in every evaluated setting, achieving up to 1.56× speedup. Ablations further show that anchor calibration is critical to generation quality. Future work will explore input-adaptive anchor scheduling in place of the fixed schedule used here.

%% file: _ref/0_style.bib
@IEEEtranBSTCTL{IEEEexample:BSTcontrol,
CTLdash_repeated_names= "no",
}


%% file: _ref/1_dit_accel.bib
@inproceedings{dit,
  author    = {William Peebles and Saining Xie},
  title     = {Scalable Diffusion Models with Transformers},
  booktitle = {Proceedings of the IEEE/CVF International Conference on Computer Vision (ICCV)},
  pages     = {4195--4205},
  year      = {2023}
}

@article{flashattn2,
  author  = {Tri Dao},
  title   = {{FlashAttention-2: Faster Attention with Better Parallelism and Work Partitioning}},
  journal = {arXiv preprint arXiv:2307.08691},
  year    = {2023}
}

@inproceedings{sageattention,
  author    = {Jintao Zhang and Jia Wei and Haofeng Huang and Pengle Zhang and Jun Zhu and Jianfei Chen},
  title     = {{SageAttention: Accurate 8-Bit Attention for Plug-and-play Inference Acceleration}},
  booktitle = {International Conference on Learning Representations (ICLR)},
  year      = {2025}
}

@article{moba,
  author  = {Enzhe Lu and Zhejun Jiang and Jingyuan Liu and Yulun Du and Tao Jiang and Chao Hong and others},
  title   = {{MoBA: Mixture of Block Attention for Long-Context LLMs}},
  journal = {arXiv preprint arXiv:2502.13189},
  year    = {2025}
}

@article{nsa,
  author  = {Jingyang Yuan and Huazuo Gao and Damai Dai and Junyu Luo and Liang Zhao and Zhengyan Zhang and others},
  title   = {{Native Sparse Attention: Hardware-Aligned and Natively Trainable Sparse Attention}},
  journal = {arXiv preprint arXiv:2502.11089},
  year    = {2025}
}

@article{h2o,
  author  = {Zhenyu Zhang and Ying Sheng and Tianyi Zhou and Tianlong Chen and Lianmin Zheng and Ruisi Cai and others},
  title   = {{H$_2$O: Heavy-Hitter Oracle for Efficient Generative Inference of Large Language Models}},
  journal = {arXiv preprint arXiv:2306.14048},
  year    = {2023}
}

@article{spatten,
  author  = {Hanrui Wang and Zhekai Zhang and Song Han},
  title   = {{SpAtten: Efficient Sparse Attention Architecture with Cascade Token and Head Pruning}},
  journal = {arXiv preprint arXiv:2012.09852},
  year    = {2020}
}

@article{pixartsigma,
  author  = {Junsong Chen and Chongjian Ge and Enze Xie and Yue Wu and Lewei Yao and Xiaozhe Ren and others},
  title   = {{PixArt-$\Sigma$: Weak-to-Strong Training of Diffusion Transformer for 4K Text-to-Image Generation}},
  journal = {arXiv preprint arXiv:2403.04692},
  year    = {2024}
}

@article{pixartalpha,
  author  = {Junsong Chen and Jincheng Yu and Chongjian Ge and Lewei Yao and Enze Xie and Yue Wu and others},
  title   = {{PixArt-$\alpha$: Fast Training of Diffusion Transformer for Photorealistic Text-to-Image Synthesis}},
  journal = {arXiv preprint arXiv:2310.00426},
  year    = {2023}
}

@article{sparsevgen2,
  author  = {Shuo Yang and Haocheng Xi and Yilong Zhao and Muyang Li and Jintao Zhang and Han Cai and others},
  title   = {{Sparse VideoGen2: Accelerate Video Generation with Sparse Attention via Semantic-Aware Permutation}},
  journal = {arXiv preprint arXiv:2505.18875},
  year    = {2025}
}

@article{sparsevdit,
  author  = {Pengtao Chen and Xianfang Zeng and Maosen Zhao and Peng Ye and Mingzhu Shen and Wei Cheng and Gang Yu and Tao Chen},
  title   = {{Sparse-vDiT: Unleashing the Power of Sparse Attention to Accelerate Video Diffusion Transformers}},
  journal = {arXiv preprint arXiv:2506.03065},
  year    = {2025}
}

@article{reattention,
  author  = {Ruichen Chen and Keith G. Mills and Liyao Jiang and Chao Gao and Di Niu},
  title   = {{Re-ttention: Ultra Sparse Visual Generation via Attention Statistical Reshape}},
  journal = {arXiv preprint arXiv:2505.22918},
  year    = {2025}
}

@article{sparsetransformer,
  author  = {Rewon Child and Scott Gray and Alec Radford and Ilya Sutskever},
  title   = {Generating Long Sequences with Sparse Transformers},
  journal = {arXiv preprint arXiv:1904.10509},
  year    = {2019}
}

@article{longformer,
  author  = {Iz Beltagy and Matthew E. Peters and Arman Cohan},
  title   = {{Longformer}: The Long-Document Transformer},
  journal = {arXiv preprint arXiv:2004.05150},
  year    = {2020}
}

@inproceedings{bigbird,
  author    = {Manzil Zaheer and Guru Guruganesh and Kumar Avinava Dubey and Joshua Ainslie and Chris Alberti and Santiago Ontanon and Philip Pham and Anirudh Ravula and Qifan Wang and Li Yang and Amr Ahmed},
  title     = {Big Bird: Transformers for Longer Sequences},
  booktitle = {Advances in Neural Information Processing Systems (NeurIPS)},
  pages     = {17283--17297},
  year      = {2020}
}

@article{reformer,
  author  = {Nikita Kitaev and Lukasz Kaiser and Anselm Levskaya},
  title   = {{Reformer}: The Efficient Transformer},
  journal = {arXiv preprint arXiv:2001.04451},
  year    = {2020}
}

@inproceedings{minference,
  author    = {Huiqiang Jiang and Yucheng Li and Chengruidong Zhang and Qianhui Wu and Xufang Luo and Surin Ahn and Zhenhua Han and Amir H. Abdi and Dongsheng Li and Chin-Yew Lin and Yuqing Yang and Lili Qiu},
  title     = {{MInference} 1.0: Accelerating Pre-filling for Long-Context {LLMs} via Dynamic Sparse Attention},
  booktitle = {Advances in Neural Information Processing Systems (NeurIPS)},
  pages     = {52481--52515},
  year      = {2024}
}

@article{teacache,
  author  = {Feng Liu and Shiwei Zhang and Xiaofeng Wang and Yujie Wei and Haonan Qiu and Yuzhong Zhao and Yingya Zhang and Qixiang Ye and Fang Wan},
  title   = {Timestep Embedding Tells: It's Time to Cache for Video Diffusion Model},
  journal = {arXiv preprint arXiv:2411.19108},
  year    = {2024}
}

@article{taylorseer,
  author  = {Jiacheng Liu and Chang Zou and Yuanhuiyi Lyu and Junjie Chen and Linfeng Zhang},
  title   = {From Reusing to Forecasting: Accelerating Diffusion Models with {TaylorSeers}},
  journal = {arXiv preprint arXiv:2503.06923},
  year    = {2025}
}

@misc{cachedit,
  author       = {Yanjun Qiu and Zhangqiu Yu and Hua Xiao and Xiuquan Hao and Dengfeng Zhang and Yunqing Li and Guiping Du and Wangpeng Zhan and Xiaoyu Zhang and Song Lin},
  title        = {{Cache-DiT}: A Unified {PyTorch}-Native Inference Engine for Diffusion Transformers},
  howpublished = {Tech report, \url{https://github.com/vipshop/cache-dit}},
  note         = {Introduces {DBCache} (Dual Block Caching); design doc: \url{https://cache-dit.readthedocs.io/en/latest/user_guide/DBCACHE_DESIGN/}},
  year         = {2025}
}

@inproceedings{distill,
  author    = {Chenlin Meng and Robin Rombach and Ruiqi Gao and Diederik Kingma and Stefano Ermon and Jonathan Ho and Tim Salimans},
  title     = {On Distillation of Guided Diffusion Models},
  booktitle = {Proceedings of the IEEE/CVF Conference on Computer Vision and Pattern Recognition (CVPR)},
  pages     = {14297--14306},
  year      = {2023}
}

@article{spargeattn,
  author  = {Jintao Zhang and Chendong Xiang and Haofeng Huang and Jia Wei and Haocheng Xi and Jun Zhu and Jianfei Chen},
  title   = {{SpargeAttn}: Accurate Sparse Attention Accelerating Any Model Inference},
  journal = {arXiv preprint arXiv:2502.18137},
  year    = {2025}
}

@article{pisa,
  author  = {Haopeng Li and Shitong Shao and Wenliang Zhong and Zikai Zhou and Lichen Bai and Hui Xiong and Zeke Xie},
  title   = {{PISA}: Piecewise Sparse Attention Is Wiser for Efficient Diffusion Transformers},
  journal = {arXiv preprint arXiv:2602.01077},
  year    = {2026}
}

@article{sparsevideogen,
  author  = {Haocheng Xi and Shuo Yang and Yilong Zhao and Chenfeng Xu and Muyang Li and Xiuyu Li and Yujun Lin and Han Cai and Jintao Zhang and Dacheng Li and Jianfei Chen and Ion Stoica and Kurt Keutzer and Song Han},
  title   = {Sparse {VideoGen}: Accelerating Video Diffusion Transformers with Spatial-Temporal Sparsity},
  journal = {arXiv preprint arXiv:2502.01776},
  year    = {2025}
}

@article{radialattn,
  author  = {Xingyang Li and Muyang Li and Tianle Cai and Haocheng Xi and Shuo Yang and Yujun Lin and Lvmin Zhang and Songlin Yang and Jinbo Hu and Kelly Peng and Maneesh Agrawala and Ion Stoica and Kurt Keutzer and Song Han},
  title   = {Radial Attention: $\mathcal{O}(n\log n)$ Sparse Attention with Energy Decay for Long Video Generation},
  journal = {arXiv preprint arXiv:2506.19852},
  year    = {2025}
}

@article{diffastattn,
  author  = {Zhihang Yuan and Hanling Zhang and Pu Lu and Xuefei Ning and Linfeng Zhang and Tianchen Zhao and Shengen Yan and Guohao Dai and Yu Wang},
  title   = {{DiTFastAttn}: Attention Compression for Diffusion Transformer Models},
  journal = {arXiv preprint arXiv:2406.08552},
  year    = {2024}
}

@article{diffastattnv2,
  author  = {Hanling Zhang and Rundong Su and Zhihang Yuan and Pengtao Chen and Mingzhu Shen and Yibo Fan and Shengen Yan and Guohao Dai and Yu Wang},
  title   = {{DiTFastAttnV2}: Head-wise Attention Compression for Multi-Modality Diffusion Transformers},
  journal = {arXiv preprint arXiv:2503.22796},
  year    = {2025}
}

@inproceedings{sd3,
  author    = {Patrick Esser and Sumith Kulal and Andreas Blattmann and Rahim Entezari and Jonas M{\"u}ller and Harry Saini and Yam Levi and Dominik Lorenz and Axel Sauer and Frederic Boesel and Dustin Podell and Tim Dockhorn and Zion English and Kyle Lacey and Alex Goodwin and Yannik Marek and Robin Rombach},
  title     = {Scaling Rectified Flow Transformers for High-Resolution Image Synthesis},
  booktitle = {Proceedings of the 41st International Conference on Machine Learning (ICML)},
  year      = {2024}
}

@article{zimage,
  author  = {{Z-Image Team} and Huanqia Cai and Sihan Cao and Ruoyi Du and Peng Gao and Aiming Hao and Steven Hoi and Zhaohui Hou and Shijie Huang and Dengyang Jiang and Yuming Jiang and Xin Jin and Liangchen Li and Zhen Li and Zhong-Yu Li and David Liu and Dongyang Liu and Qilong Wu and Feng Yu and Zechao Zhan and Chi Zhang and Shifeng Zhang and Ruikai Zhou and Shilin Zhou},
  title   = {{Z-Image}: An Efficient Image Generation Foundation Model with Single-Stream Diffusion Transformer},
  journal = {arXiv preprint arXiv:2511.22699},
  year    = {2025}
}

@inproceedings{flashattn,
  author    = {Tri Dao and Daniel Y. Fu and Stefano Ermon and Atri Rudra and Christopher R{\'e}},
  title     = {{FlashAttention}: Fast and Memory-Efficient Exact Attention with {IO}-Awareness},
  booktitle = {Advances in Neural Information Processing Systems (NeurIPS)},
  year      = {2022}
}

@inproceedings{quest,
  author    = {Jiaming Tang and Yilong Zhao and Kan Zhu and Guangxuan Xiao and Baris Kasikci and Song Han},
  title     = {{QUEST}: Query-Aware Sparsity for Efficient Long-Context {LLM} Inference},
  booktitle = {Proceedings of the 41st International Conference on Machine Learning (ICML)},
  pages     = {47901--47911},
  year      = {2024}
}

@article{xattention,
  author  = {Ruyi Xu and Guangxuan Xiao and Haofeng Huang and Junxian Guo and Song Han},
  title   = {{XAttention}: Block Sparse Attention with Antidiagonal Scoring},
  journal = {arXiv preprint arXiv:2503.16428},
  year    = {2025}
}

@article{seerattention,
  author  = {Yizhao Gao and Zhichen Zeng and Dayou Du and Shijie Cao and Peiyuan Zhou and Jiaxing Qi and Junjie Lai and Hayden Kwok-Hay So and Ting Cao and Fan Yang and Mao Yang},
  title   = {{SeerAttention}: Learning Intrinsic Sparse Attention in Your {LLMs}},
  journal = {arXiv preprint arXiv:2410.13276},
  year    = {2024}
}

@misc{flux2,
  author       = {{Black Forest Labs}},
  title        = {{FLUX.2}},
  howpublished = {\url{https://bfl.ai/models/flux-2}},
  note         = {Official inference repository: \url{https://github.com/black-forest-labs/flux2}},
  year         = {2025}
}

@article{vsa,
  author  = {Peiyuan Zhang and Yongqi Chen and Haofeng Huang and Will Lin and Zhengzhong Liu and Ion Stoica and Eric Xing and Hao Zhang},
  title   = {{VSA: Faster Video Diffusion with Trainable Sparse Attention}},
  journal = {arXiv preprint arXiv:2505.13389},
  year    = {2025}
}

@article{vorta,
  author  = {Wenhao Sun and Rong-Cheng Tu and Yifu Ding and Zhao Jin and Jingyi Liao and Shunyu Liu and Dacheng Tao},
  title   = {{VORTA: Efficient Video Diffusion via Routing Sparse Attention}},
  journal = {arXiv preprint arXiv:2505.18809},
  year    = {2025}
}

@article{sla,
  author  = {Jintao Zhang and Haoxu Wang and Kai Jiang and Shuo Yang and Kaiwen Zheng and Haocheng Xi and others},
  title   = {{SLA: Beyond Sparsity in Diffusion Transformers via Fine-Tunable Sparse-Linear Attention}},
  journal = {arXiv preprint arXiv:2509.24006},
  year    = {2025}
}

@article{gsa,
  author  = {Sucheng Ren and Qihang Yu and Ju He and Alan Yuille and Liang-Chieh Chen},
  title   = {{Grouping First, Attending Smartly: Training-Free Acceleration for Diffusion Transformers}},
  journal = {arXiv preprint arXiv:2505.14687},
  year    = {2025}
}

@article{dsv,
  author  = {Xin Tan and Yuetao Chen and Yimin Jiang and Xing Chen and Kun Yan and Nan Duan and others},
  title   = {{DSV: Exploiting Dynamic Sparsity to Accelerate Large-Scale Video DiT Training}},
  journal = {arXiv preprint arXiv:2502.07590},
  year    = {2025}
}

@article{ultraimagegen,
  author  = {Yuyao Zhang and Yu-Wing Tai},
  title   = {{UltraImageGen: Efficient Ultra-High-Resolution Image Generation with Hierarchical Local Attention}},
  journal = {arXiv preprint arXiv:2510.16325},
  year    = {2025}
}

@article{spargeattn2,
  author  = {Jintao Zhang and Kai Jiang and Chendong Xiang and Weiqi Feng and Yuezhou Hu and Haocheng Xi and Jianfei Chen and Jun Zhu},
  title   = {{SpargeAttention2: Trainable Sparse Attention via Hybrid Top-k+Top-p Masking and Distillation Fine-Tuning}},
  journal = {arXiv preprint arXiv:2602.13515},
  year    = {2026}
}

@article{rainfusion2,
  author  = {Aiyue Chen and Yaofu Liu and Junjian Huang and Guang Lian and Yiwu Yao and Wangli Lan and others},
  title   = {{RainFusion2.0: Temporal-Spatial Awareness and Hardware-Efficient Block-wise Sparse Attention}},
  journal = {arXiv preprint arXiv:2512.24086},
  year    = {2025}
}

@article{liu2024deepseek,
  title={Deepseek-v3 technical report},
  author={Liu, Aixin and Feng, Bei and Xue, Bing and Wang, Bingxuan and Wu, Bochao and Lu, Chengda and Zhao, Chenggang and Deng, Chengqi and Zhang, Chenyu and Ruan, Chong and others},
  journal={arXiv preprint arXiv:2412.19437},
  year={2024}
}

@article{bai2023qwen,
  title={Qwen technical report},
  author={Bai, Jinze and Bai, Shuai and Chu, Yunfei and Cui, Zeyu and Dang, Kai and Deng, Xiaodong and Fan, Yang and Ge, Wenbin and Han, Yu and Huang, Fei and others},
  journal={arXiv preprint arXiv:2309.16609},
  year={2023}
}


%% file: _ref/2_dataset.bib
@inproceedings{geneval,
  author    = {Dhruba Ghosh and Hanna Hajishirzi and Ludwig Schmidt},
  title     = {{GenEval}: An Object-Focused Framework for Evaluating Text-to-Image Alignment},
  booktitle = {Advances in Neural Information Processing Systems (NeurIPS), Datasets and Benchmarks Track},
  year      = {2023}
}

@article{hpsv2,
  author  = {Xiaoshi Wu and Yiming Hao and Keqiang Sun and Yixiong Chen and Feng Zhu and Rui Zhao and Hongsheng Li},
  title   = {Human Preference Score v2: A Solid Benchmark for Evaluating Human Preferences of Text-to-Image Synthesis},
  journal = {arXiv preprint arXiv:2306.09341},
  year    = {2023}
}

@inproceedings{imagereward,
  author    = {Jiazheng Xu and Xiao Liu and Yuchen Wu and Yuxuan Tong and Qinkai Li and Ming Ding and Jie Tang and Yuxiao Dong},
  title     = {{ImageReward}: Learning and Evaluating Human Preferences for Text-to-Image Generation},
  booktitle = {Advances in Neural Information Processing Systems (NeurIPS)},
  year      = {2023}
}

@article{mjhq30k,
  author  = {Daiqing Li and Aleks Kamko and Ehsan Akhgari and Ali Sabet and Linmiao Xu and Suhail Doshi},
  title   = {Playground v2.5: Three Insights towards Enhancing Aesthetic Quality in Text-to-Image Generation},
  journal = {arXiv preprint arXiv:2402.17245},
  year    = {2024}
}

@inproceedings{fid,
  author    = {Martin Heusel and Hubert Ramsauer and Thomas Unterthiner and Bernhard Nessler and Sepp Hochreiter},
  title     = {{GANs} Trained by a Two Time-Scale Update Rule Converge to a Local {Nash} Equilibrium},
  booktitle = {Advances in Neural Information Processing Systems (NeurIPS)},
  year      = {2017}
}

@inproceedings{clipscore,
  author    = {Jack Hessel and Ari Holtzman and Maxwell Forbes and Ronan Le Bras and Yejin Choi},
  title     = {{CLIPScore}: A Reference-free Evaluation Metric for Image Captioning},
  booktitle = {Proceedings of the 2021 Conference on Empirical Methods in Natural Language Processing (EMNLP)},
  year      = {2021}
}


%% file: _ref/3_mobile.bib
@article{heterollm,
  author  = {Le Chen and Dahu Feng and Erhu Feng and Rong Zhao and Yingrui Wang and Yubin Xia and Haibo Chen and Pinjie Xu},
  title   = {{HeteroLLM}: Accelerating Large Language Model Inference on Mobile {SoCs} with Heterogeneous {AI} Accelerators},
  journal = {arXiv preprint arXiv:2501.14794},
  year    = {2025}
}

@article{llmnpu,
  author  = {Daliang Xu and Hao Zhang and Liming Yang and Ruiqi Liu and Gang Huang and Mengwei Xu and Xuanzhe Liu},
  title   = {Fast On-device {LLM} Inference with {NPUs}},
  journal = {arXiv preprint arXiv:2407.05858},
  year    = {2024}
}

@article{powerinfer2,
  author  = {Zhenliang Xue and Yixin Song and Zeyu Mi and Xinrui Zheng and Yubin Xia and Haibo Chen},
  title   = {{PowerInfer-2}: Fast Large Language Model Inference on a Smartphone},
  journal = {arXiv preprint arXiv:2406.06282},
  year    = {2024}
}

@article{shadowattn,
  author  = {Wangsong Yin and Daliang Xu and Mengwei Xu and Gang Huang and Xuanzhe Liu},
  title   = {Dynamic Sparse Attention on Mobile {SoCs}},
  journal = {arXiv preprint arXiv:2508.16703},
  year    = {2025}
}

@misc{sdcpp,
  author       = {{leejet}},
  title        = {{stable-diffusion.cpp}: Stable Diffusion and {Flux} in pure {C/C++}},
  howpublished = {\url{https://github.com/leejet/stable-diffusion.cpp}},
  year         = {2024}
}

@misc{ggml,
  author       = {{ggml-org}},
  title        = {{ggml}: Tensor Library for Machine Learning},
  howpublished = {\url{https://github.com/ggml-org/ggml}},
  year         = {2022}
}

@misc{fastrpc,
  author       = {{Qualcomm Technologies, Inc.}},
  title        = {{FastRPC}: Userspace Library for Remote Procedure Calls between the {CPU} and the {DSP}},
  howpublished = {\url{https://github.com/qualcomm/fastrpc}},
  year         = {2024}
}

@inproceedings{llvm,
  author    = {Chris Lattner and Vikram Adve},
  title     = {{LLVM}: A Compilation Framework for Lifelong Program Analysis and Transformation},
  booktitle = {Proceedings of the International Symposium on Code Generation and Optimization (CGO)},
  pages     = {75--86},
  year      = {2004},
  doi       = {10.1109/CGO.2004.1281665}
}

@misc{hexagonsdk,
  author       = {{Qualcomm Technologies, Inc.}},
  title        = {{Qualcomm} {Hexagon} {NPU} {SDK}},
  howpublished = {\url{https://www.qualcomm.com/developer/software/hexagon-npu-sdk}},
  year         = {2025}
}

@misc{clqcomionhostptr,
  author       = {{Khronos Group}},
  title        = {{OpenCL} Extension: {\texttt{cl\_qcom\_ion\_host\_ptr}}},
  howpublished = {\url{https://registry.khronos.org/OpenCL/extensions/qcom/cl_qcom_ion_host_ptr.txt}},
  note         = {Version 7},
  year         = {2018}
}

@misc{htpopslib,
  author       = {{haozixu}},
  title        = {{htp-ops-lib}: Self-implemented NN Operators for {Qualcomm}'s {Hexagon} {NPU}},
  howpublished = {\url{https://github.com/haozixu/htp-ops-lib}},
  year         = {2025}
}

@inproceedings{hexagonnpu,
  author    = {Eric Mahurin},
  title     = {{Qualcomm} {Hexagon} {NPU}},
  booktitle = {2023 IEEE Hot Chips 35 Symposium (HCS)},
  pages     = {1--19},
  year      = {2023},
  doi       = {10.1109/HCS59251.2023.10254715},
  note      = {Hexagon NPU architecture and on-chip memory}
}

@misc{qnn,
  author       = {{Qualcomm Technologies, Inc.}},
  title        = {{Qualcomm AI Engine Direct SDK} ({QNN})},
  howpublished = {\url{https://www.qualcomm.com/developer/software/qualcomm-ai-engine-direct-sdk}},
  year         = {2024}
}

@misc{dmabuf,
  author       = {{The Linux Kernel Community}},
  title        = {Buffer Sharing and Synchronization ({dma-buf})},
  howpublished = {Linux Kernel Documentation, \url{https://docs.kernel.org/driver-api/dma-buf.html}},
  year         = {2026}
}

@misc{opencl,
  author       = {{Khronos Group}},
  title        = {The {OpenCL} Specification},
  howpublished = {\url{https://registry.khronos.org/OpenCL/specs/3.0-unified/html/OpenCL_API.html}},
  note         = {Version 3.0},
  year         = {2023}
}
